\documentclass{article}

\PassOptionsToPackage{table}{xcolor}
\usepackage{PRIMEarxiv}

\usepackage[utf8]{inputenc} 
\usepackage[T1]{fontenc}    
\usepackage{hyperref}       
\usepackage{url}            
\usepackage{booktabs}       
\usepackage{amsfonts}       
\usepackage{nicefrac}       
\usepackage{microtype}      
\usepackage{lipsum}
\usepackage{fancyhdr}       
\usepackage{graphicx}       
\graphicspath{{media/}}     

\usepackage{amsmath,amsfonts}
\usepackage{algorithmic}
\usepackage{algorithm}
\usepackage{array}
\usepackage{textcomp}
\usepackage{stfloats}
\usepackage{tcolorbox}
\usepackage{url}
\usepackage{verbatim}
\usepackage{graphicx}
\usepackage{cite}
\usepackage{amssymb}
\usepackage{multirow}
\usepackage{ulem}
\usepackage{listings}
\usepackage{booktabs}
\usepackage{makecell}
\usepackage{threeparttable}
\usepackage{cleveref}
\title{RingMo-Agent: A Unified Remote Sensing Foundation Model for Multi-Platform and Multi-Modal Reasoning}

\author{
  \textbf{Huiyang Hu, Peijin Wang, Yingchao Feng, Kaiwen Wei, Wenxin Yin, Wenhui Diao, Mengyu Wang,} \\
  \textbf{Hanbo Bi, Kaiyue Kang, Tong Ling,  Kun Fu, Xian Sun} \\
  Aerospace Information Research Institute, Chinese Academy of Sciences\\
School of Electronic, Electrical and Communication Engineering, University of Chinese Academy of Sciences\\
University of Chinese Academy of Sciences\\
Key Laboratory of Target Cognition and Application Technology (TCAT)\\
  \texttt{\{huhuiyang22, wangpeijin17\}mails.ucas.ac.cn} \\
}

\begin{document}
\maketitle
 \begin{abstract}
Remote sensing (RS) images from multiple modalities and platforms exhibit diverse details due to differences in sensor characteristics and imaging perspectives. Existing vision-language research in RS largely relies on relatively homogeneous data sources. Moreover, they still remain limited to conventional visual perception tasks such as classification or captioning. As a result, these methods fail to serve as a unified and standalone framework capable of effectively handling RS imagery from diverse sources in real-world applications. To address these issues, we propose RingMo-Agent, a model designed to handle multi-modal and multi-platform data that performs perception and reasoning tasks based on user textual instructions.  Compared with existing models, RingMo-Agent 1) is supported by a large-scale vision-language  dataset named RS-VL3M, comprising over 3 million image-text pairs, spanning optical, SAR, and infrared (IR) modalities collected from both satellite and UAV platforms, covering perception and challenging reasoning tasks; 2) learns modality adaptive representations by incorporating separated embedding layers to construct isolated features for heterogeneous modalities and reduce cross-modal interference; 3) functions as an agent-based framework equipped with external tool-use capabilities, employing a reinforcement learning-optimized trajectory decoding mechanism that integrates visual grounding tools for long-horizon embodied navigation tasks. Extensive experiments on various RS vision-language tasks demonstrate that RingMo-Agent not only proves effective in both visual understanding and sophisticated analytical tasks, but also exhibits strong generalizability across different platforms and sensing modalities.

\end{abstract}


\section{Introduction}
With  the remarkable advancements of large language models (LLMs) in semantic understanding and reasoning, vision-language models designed for open-world environments have experienced rapid development \cite{10938647, 10970423,11086426}. \textcolor{black}{Leveraging the capabilities of advanced LLMs such as DeepSeek \cite{guo2025deepseek,dai2024deepseekmoe, liu2024deepseek,deepseekai2025deepseekv3technicalreport}, GPT \cite{brown2020language, achiam2023gpt},  and Llama \cite{touvron2023llama,touvron2023llama2, dubey2024llama}, these models are now capable of perceiving complex semantics, conducting multi-turn interactions, and performing context-aware task planning. This evolution marks a significant shift from static perception toward dynamic understanding and autonomous decision-making, thereby enabling broader applications in intelligent interaction and high-level scene reasoning.}

\begin{figure}[tbp]
    \centering
    \includegraphics[width=0.5\linewidth]{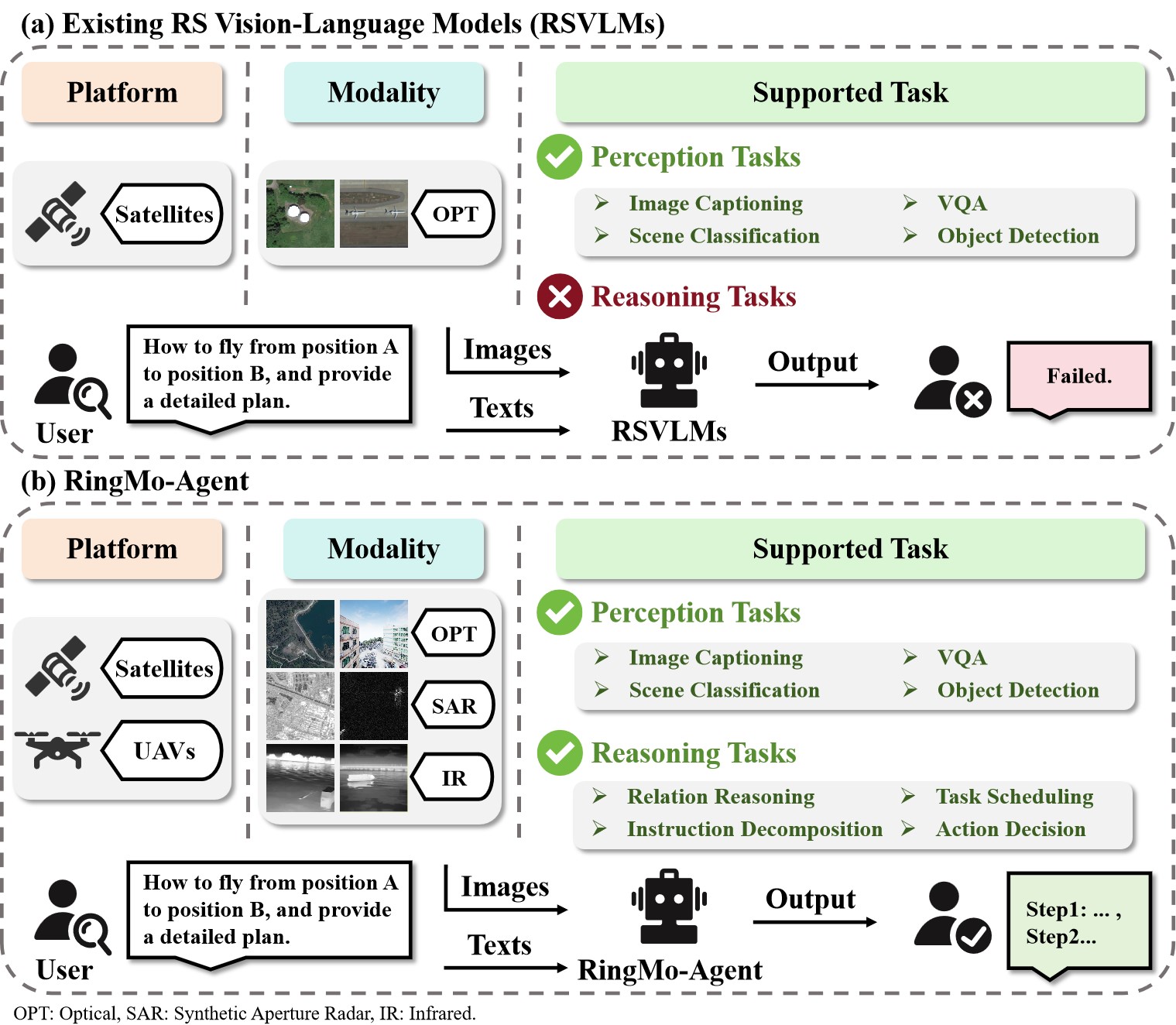}
    \caption{Comparison of RingMo-Agent and most existing RS vision-language models in terms of data platforms, modalities, and supported tasks.}
\label{fig:comparison}
\end{figure}

Driven by recent advances in these research for natural scenes, the paradigm of vision-language models has gradually extended to RS \cite{li2024vision}, introducing capabilities such as instruction tuning and joint vision-language modeling into RS image analysis. These efforts have led to the emergence of models that demonstrate promising performance in fundamental RS perception tasks such as object recognition and scene classification \cite{zhang2024earthgpt, zhang2024popeye, silva2024large,muhtar2024lhrs, zhan2025skyeyegpt, luo2024skysensegpt, kuckreja2024geochat, pang2024h2rsvlm, wang2024ringmogpt, hu2023rsgpt, bazi2024rs}. However, current vision-language models for RS still suffer from significant limitations:

\begin{itemize}
\item \textbf{Limitations of single-source modeling in diverse RS data.} Due to variations in data acquisition platforms (e.g., satellites and UAVs) and imaging modalities (e.g., optical, SAR, and infrared), RS data exhibit significant differences in physical characteristics such as spatial resolution, viewing geometry, and spectral response. These heterogeneous properties fundamentally affect the way information is represented in the imagery. However, most existing vision-language models for RS are trained on data from a single modality or platform, limiting their ability to handle the complexity of multi-source data in real-world scenarios. Consequently, these approaches often suffer from poor generalization and limited compatibility in cross-platform and cross-modal applications.
    
\item \textbf{Limitations of existing reasoning task paradigms.} 
Existing RS vision-language research  primarily focuses on specific visual perception tasks.  However, these tasks remain confined to foundational perception and recognition, with models lacking training in advanced cognitive abilities such as relation reasoning among scene entities and action prediction. Therefore, these models are restricted to narrow applications and struggle to replicate the superior intelligence demonstrated by agent systems in natural scenes when faced with high-level tasks requiring decision-making,  as illustrated in \cref{fig:comparison}. This limitation fundamentally stems from the lack of large-scale RS image-text datasets oriented toward complex semantic interactions and reasoning, which restricts the models' ability to generalize multi-modal knowledge.
\end{itemize}

\begin{table*}[htbp]
\centering
\caption{Comparison of model architectures and capabilities, including LLMs, visual encoders, supported modalities, and task coverage. SAR: Synthetic Aperture Radar, IR: Infrared, IC: Image Captioning, CL: Classification, RR: Relation Reasoning, OD: Object Detection (including visual grounding), ID: Instruction Decomposition, AD: Action Decision, TS: Task Scheduling.}
\label{tab:models}
\renewcommand{\arraystretch}{1.2}
\resizebox{1.0\textwidth}{!}{%
\begin{tabular}{lccc|ccc|cccc|cccc}
\toprule
\multicolumn{1}{l}{\multirow{2}{*}{Model}} & \multicolumn{1}{c}{\multirow{2}{*}{LLM}} & \multicolumn{1}{c}{\multirow{2}{*}{Visual Encoder}} & \multicolumn{1}{c|}{\multirow{2}{*}{Size}} & \multicolumn{3}{c|}{Modalities} & \multicolumn{4}{c|}{Perception Tasks} & \multicolumn{4}{c}{Reasoning Tasks}  \\ \cline{5-15} 
\multicolumn{1}{c}{} & \multicolumn{1}{c}{} & \multicolumn{1}{c}{}  & \multicolumn{1}{c|}{}  & Optical & SAR & IR & IC   &VQA &CL & OD  & RR & ID & AD & TS \\ \midrule
EarthGPT \cite{zhang2024earthgpt} & LLaMA-2 & \makecell{DINOv2 ViT-L/14, \\ CLIP ConvNeXt-L} & - & \checkmark& \checkmark& \checkmark & \checkmark& \checkmark& \checkmark& \checkmark  &&& &\\
Popeye \cite{zhang2024popeye} & LLaMA-2 (7B) &\makecell{DINOv2 ViT-L/14, \\ CLIP ViT-L/14} & - & \checkmark& \checkmark & &   & &&\checkmark &&  \\
   RS-CapRet \cite{silva2024large} & LLaMA-2 (7B) & CLIP ViT-L/14 &224 & \checkmark && & \checkmark &&&& &&&\\
   LHRS-Bot\cite{muhtar2024lhrs} & LLaMA-2 (7B) & CLIP ViT-L/14 &224 & \checkmark & & & \checkmark& \checkmark & \checkmark & \checkmark &&& & \\
   SkyEyeGPT \cite{zhan2025skyeyegpt} & LLaMA-2-Chat (7B) & EVA ViT-G  & 448 & \checkmark & & & \checkmark & \checkmark &\checkmark & \checkmark & &&  & \\ 
SkySenseGPT \cite{luo2024skysensegpt} & Vicuna-v1.5   & CLIP ViT-L/14  & 504 & \checkmark  &   & & \checkmark & \checkmark & \checkmark  & \checkmark & \checkmark &   & & \\
   GeoChat  \cite{kuckreja2024geochat} & Vicuna-v1.5 (7B) & CLIP ViT-L/14& 504 & \checkmark & & & \checkmark & \checkmark & \checkmark  & \checkmark & && &\\
   H$^2$RSVLM \cite{pang2024h2rsvlm}    & Vicuna-v1.5 (7B) &CLIP ViT-L/14 &336 & \checkmark & & & \checkmark & \checkmark& \checkmark  & \checkmark &&&& \\
RingMoGPT \cite{wang2024ringmogpt}  & Vicuna-v1.1 (13B)  & EVA ViT-G  & 448& \checkmark  &   & & \checkmark   & \checkmark & \checkmark  & \checkmark  &&&&    \\
RSGPT \cite{hu2023rsgpt} & Vicuna (7B/13B) & EVA ViT-G &224   & \checkmark & & & \checkmark & \checkmark &&&&&&\\
RS-LLaVA \cite{bazi2024rs} & Vicuna-v1.5 (7B/13B)  & CLIP ViT-L/14 &336  & \checkmark & & & \checkmark  & \checkmark  &&&&&&\\
 \cellcolor{gray!10}\textbf{RingMo-Agent (ours)}     &\cellcolor{gray!10}\textbf{DeepSeekMoE (3B)} &\cellcolor{gray!10}\textbf{SigLIP-SO400M-384}  &\cellcolor{gray!10}\textbf{Any} &\cellcolor{gray!10}\checkmark &\cellcolor{gray!10}\checkmark &\cellcolor{gray!10}\checkmark &\cellcolor{gray!10}\checkmark &\cellcolor{gray!10}\checkmark &\cellcolor{gray!10}\checkmark &\cellcolor{gray!10}\checkmark &\cellcolor{gray!10}\checkmark &\cellcolor{gray!10}\checkmark &\cellcolor{gray!10}\checkmark &\cellcolor{gray!10}\checkmark
     \\ \bottomrule
\end{tabular}%
}
\end{table*}

\begin{figure*}[t]
    \centering
    \includegraphics[width=0.9\textwidth]{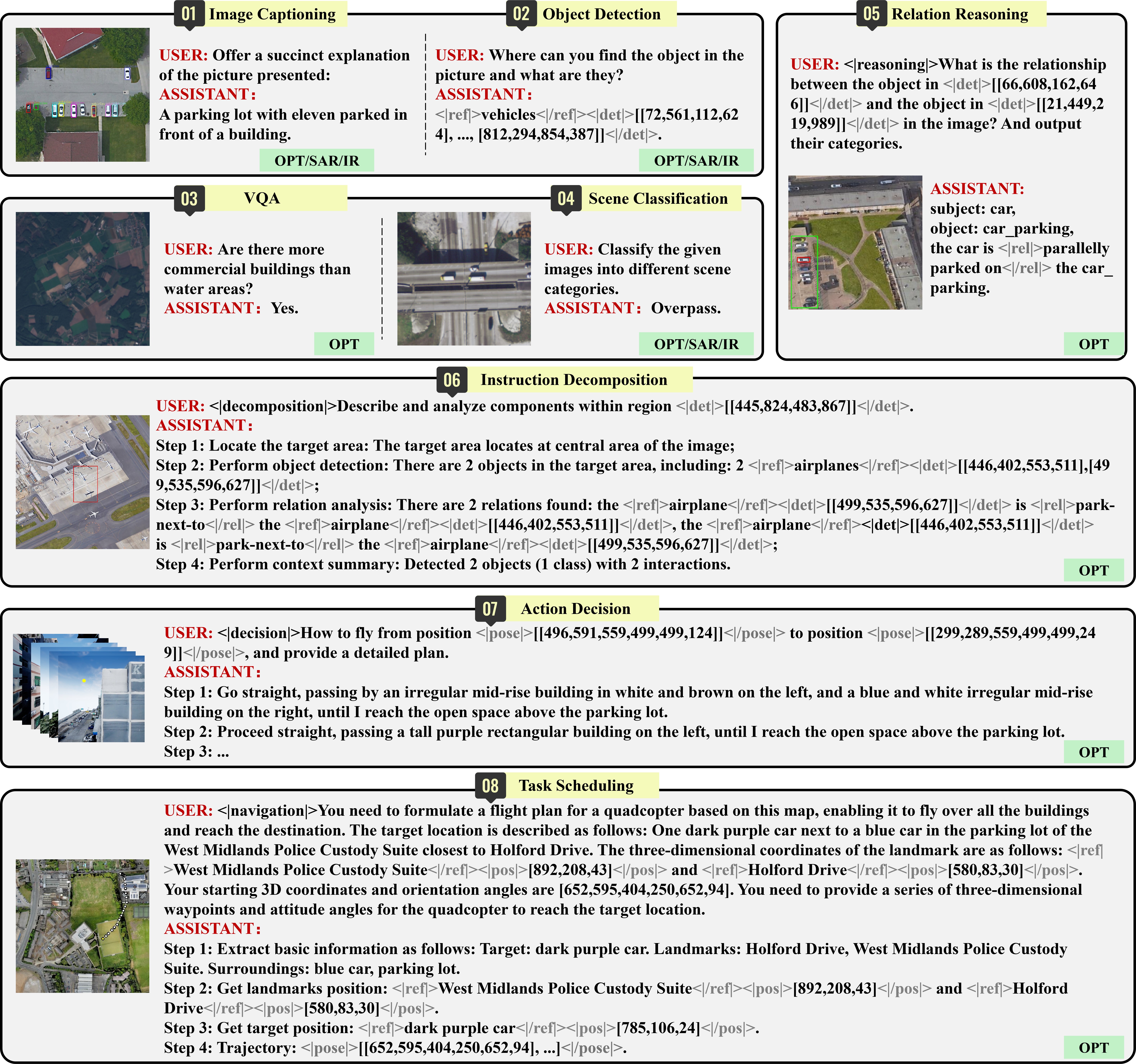}
    \caption{RingMo-Agent supports perception tasks such as image captioning, object detection, VQA, and scene classification, and reasoning tasks including relation reasoning, instruction decomposition, action decision, and task scheduling. Some tasks are enabled by optical (OPT), SAR, and infrared (IR) modalities.}
\label{fig:task}
\end{figure*}

To address the aforementioned limitations, we propose RingMo-Agent, a model designed to handle multi-modal and multi-platform data, capable of performing perception and reasoning tasks based on user instructions. \textbf{First}, to unleash the potential of vision-language models in intelligent agent applications, we construct a vision-language dataset named RS-VL3M, comprising 3 million image-text pairs across three modalities (optical, SAR, and infrared), two platforms (satellites and UAVs), and eight tasks, as shown in \cref{fig:task}. \textbf{Second}, \textcolor{black}{we equip RingMo-Agent with external tool-use capabilities for complex spatial navigation. Specifically, it invokes an external visual grounding tool to obtain absolute target coordinates, which guide an internal decoder to autoregressively predict spatial trajectories. This process is further optimized with a Group Relative Policy Optimization (GRPO)-based reinforcement learning (RL) stage to reduce exposure bias and improve navigation robustness.} \textbf{Third}, a modality-aware visual encoder with separated embeddings is incorporated to mitigate distribution shifts across different sensing platforms and modalities, supporting robust feature extraction and alignment. \textbf{Finally}, unlike previous methods that mainly focus on optical imagery and fundamental perception tasks, RingMo-Agent unifies both perception and reasoning across heterogeneous modalities and platforms, as summarized in \Cref{tab:models}. We further conduct comprehensive comparisons between our proposed model and existing advanced methods on public and self-constructed multi-source RS datasets. 

The main contributions can be summarized as follows: 

\begin{enumerate} 
    \item We propose RingMo-Agent, a model that supports three sensing modalities and eight task types across two platforms, enabling a unified framework spanning from basic visual perception to \textcolor{black}{advanced spatial reasoning and embodied navigation.} 
    \item We construct RS-VL3M, the first large-scale RS vision-language dataset integrating multi-platform and multi-modal data, with over 3 million image-text pairs. It also includes a high-quality shared multi-modal subset specifically designed for multi-turn dialogue evaluation. 
    \item \textcolor{black}{RingMo-Agent adopts modality-specific embedding layers to reduce distribution shifts across heterogeneous modalities, and further introduces a RL-enhanced trajectory decoder coupled with external visual grounding tools, enabling explicit goal-directed navigation over long horizons.}
\end{enumerate}

Under the validation, RingMo-Agent shows strong performance across eight RS tasks, outperforming both expert models and generalist models, while exhibiting robust zero-shot generalization. 

\section{Related works}

\subsection{Vision-Language Research in Natural Scenes}

\textcolor{black}{In recent years, vision-language models have made significant advancements, becoming an essential bridge between visual and language understanding.} 

\textcolor{black}{Early studies primarily focused on perception-level tasks such as image-text matching, image captioning, and closed-form question answering, emphasizing cross-modal alignment and representation learning between vision and language. For instance, BLIP-2 \cite{li2023blip} employs three loss strategies to align visual and textual content, achieving strong performance on VQA and image-text retrieval. InstructBLIP \cite{dai2023instructblip} explores a unified instruction-tuning paradigm, where instruction-aware visual feature extraction is employed to accomplish fundamental visual perception tasks. Later works investigated pixel-level tasks including localization, detection, and segmentation. MiniGPT-v2 \cite{chen2023minigpt}, Kosmos-2 \cite{peng2023kosmos}, and Shikra \cite{chen2023shikra} explore representing spatial coordinates via natural language. LISA \cite{lai2024lisa} further introduces an embedding-as-mask paradigm, decoding segmentation masks by predicting the hidden representation of a special token.}

\textcolor{black}{With the advancement of model capabilities, research exemplified by GPT-4 \cite{achiam2023gpt} has gradually expanded to encompass open-ended question answering, visual commonsense reasoning, and multi-turn dialogue tasks, incorporating contextual modeling and knowledge reasoning. The Gemini family \cite{team2023gemini, team2024gemini, comanici2025gemini} provides multiple model variants tailored to different computational constraints and application scenarios, demonstrating state-of-the-art performance across multiple dimensions including multi-modal understanding, visual aesthetics, and tool utilization. GPT-4o \cite{hurst2024gpt} exhibits superior performance across 50 languages, with substantially enhanced capabilities for processing non-English text. Its lightweight counterpart, GPT-4o Mini \cite{2025gpt4omini}, offers a more balanced trade-off between performance and computational cost. The Qwen-VL series \cite{bai2023qwenvlversatilevisionlanguagemodel, wang2024qwen2, bai2025qwen3vltechnicalreport} achieves highly competitive results on benchmarks covering coding, mathematics, and general reasoning. Leveraging the low-cost and high-performance characteristics of the DeepSeekMoE \cite{dai2024deepseekmoe, wang2024auxiliary}, DeepSeek-VL2 \cite{wu2024deepseek}, trained on diverse multi-modal data, is capable of accurately interpreting scientific plots and analyzing various Internet memes.} 

\textcolor{black}{Recent studies have highlighted autonomous agents built upon LLMs as a promising research direction for achieving human-level decision making, where agent architectures are designed to perceive dynamic environments and translate planning outcomes into executable actions \cite{wang2024survey}. For example, Reflexion \cite{shinn2023reflexion} introduces a self-reflection mechanism that enables performance improvement without fine-tuning LLMs, thereby maintaining low computational overhead. HuggingGPT \cite{shen2023hugginggpt} proposes integrating a rich and powerful ecosystem of community-developed AI models, where tasks are planned and appropriate models are selected to execute subtasks and generate responses. Toolformer \cite{schick2023toolformer}, on the other hand, explores the ability of LLMs to make decisions and autonomously invoke external tools via API interfaces. MetaGPT \cite{hong2023metagpt} adopts an assembly line paradigm to organize multiple agents for collaborative cooperation.}

Current models are evolving from perception-driven systems to generalist agents with reasoning capabilities, but remain largely focused on natural scenes and struggle with the complexity of RS imagery.

\subsection{Vision-Language Research in Remote Sensing}

As RS foundation models advance \cite{hu2025rs,11298536}, integrating LLMs has become an emerging direction. However, due to data heterogeneity and task diversity, their application in RS remains largely exploratory, with most methods relying on fine-tuning existing architectures.

\textcolor{black}{With works such as RSGPT \cite{hu2023rsgpt} and RSVG \cite{zhan2023rsvg} demonstrating that vision-language models can generalize effectively on RS images, research has increasingly focused on the design and training of these models. RemoteCLIP \cite{liu2024remoteclip} establishes a unified RS vision-language model and validates its effectiveness across downstream tasks such as image classification, image-text retrieval, and object counting. GeoChat \cite{kuckreja2024geochat} introduces a RS vision-language model with multi-task dialogue capabilities, primarily focusing on object existence, as well as attribute judgments such as color and quantity. It also makes preliminary attempts to explore the relationship between object presence and scene semantics. SkyEyeGPT \cite{zhan2025skyeyegpt} incorporates a UAV video captioning task and investigates the ability of a RS foundation model to jointly learn from both satellite and UAV imagery.  RingMoGPT \cite{wang2024ringmogpt} further extend the task scope to include change detection and grounded captioning.}

However, existing studies mainly focus on perception tasks using optical imagery, covering only a limited portion of RS data.  Recently, EarthGPT \cite{zhang2024earthgpt} incorporates SAR and infrared data to enhance the multi-modal detection capabilities of vision-language models. SkySenseGPT \cite{luo2024skysensegpt} explore the model’s ability to reason about relationships between objects in images. AeroVerse \cite{yao2024aeroverse} attempts to integrate UAV-based intelligent agent tasks into the vision-language model to explore capabilities in navigation exploration. 

Despite recent progress, current RS research remains largely confined to basic perception representation, lacking autonomous decision-making capabilities. Moreover, they predominantly focus on single-platform and single-modal data, which limits their applicability in real-world RS scenarios.

\subsection{Tasks and Datasets of Remote Sensing Vision-Language Research}

To support the training of RS intelligent systems, numerous vision-language datasets have been developed, most of which are extended from single-modality datasets originally designed for tasks such as object detection, semantic segmentation, image classification, and change detection, and have gradually evolved into large-scale and multi-modal  benchmarks.

Early datasets were mostly based on manually annotated image descriptions, with limited scale semantic levels, mainly  supporting basic tasks such as VQA and classification, as exemplified by RSVQA \cite{lobry2020rsvqa} and AID \cite{xia2017aid}. With the expansion of tasks, dataset types and scales have gradually grown. For instance,  RemoteCLIP \cite{liu2024remoteclip} constructs a large-scale caption dataset by applying rule-based methods to RS datasets designed for detection, segmentation, and retrieval, aligning visual and linguistic data. RSVG \cite{zhan2023rsvg}, built upon the optical object detection dataset DIOR \cite{li2020object}, conducts attribute analysis on image objects and generates a visual grounding dataset through rule-based processing. RSGPT \cite{hu2023rsgpt} extends the optical object detection dataset DOTA \cite{xia2018dota}, leveraging expert annotations from RS specialists to develop a comprehensive image captioning dataset.

\textcolor{black}{Several studies have aimed to build large-scale, high-quality RS image-text datasets encompassing diverse task types. XLRS-Bench \cite{wang2025xlrs} is designed for optical ultra-high-resolution RS imagery and provides a comprehensive evaluation benchmark encompassing 16 sub-tasks across perception and reasoning. VRSBench \cite{li2024vrsbench} provides a large-scale RS vision-language dataset comprising 29,614 images, with annotations covering three key application scenarios: object referring, VQA, and human-verified detailed captions. EarthGPT \cite{zhang2024earthgpt} collects a diverse range of RS data covering five task types and three image modalities, designing different instructions to construct a pretraining dataset. Additionally, models such as GeoChat \cite{kuckreja2024geochat}, RingMoGPT \cite{wang2024ringmogpt}, and SkySenseGPT \cite{luo2024skysensegpt} leverage constrained LLMs to generate responses for augmented tasks, serving as data for training.}

Despite these advancements, there remains a lack of datasets capable of supporting diverse task types, multiple modalities, and various platforms, which is essential for advancing toward generalist intelligent agents.

\section{Dataset Construction}

\subsection{Overall Data Analysis}

\textcolor{black}{We construct a large-scale, high-quality RS image-text dataset named RS-VL3M, encompassing images captured by various sensors mounted on different imaging platforms. The dataset features images with diverse resolutions, viewing angles, and imaging mechanisms, totaling over 3 million image-text pairs. It supports two major categories of tasks: reasoning-oriented tasks (task scheduling, action decision, instruction decomposition, and relation reasoning), and perception-oriented tasks (image captioning, object detection, image classification, and VQA), as detailed in \cref{tab1}. Furthermore, to enhance the multi-turn dialogue capability of the model under consistent data conditions, we establish a multi-modal dialogue subset based on a shared image pool. This subset enables interactive classification, detection, and captioning tasks on infrared and SAR modalities simultaneously for the same images.}

\begin{figure*}[tbp]
    \centering
    \includegraphics[width=0.9\textwidth]{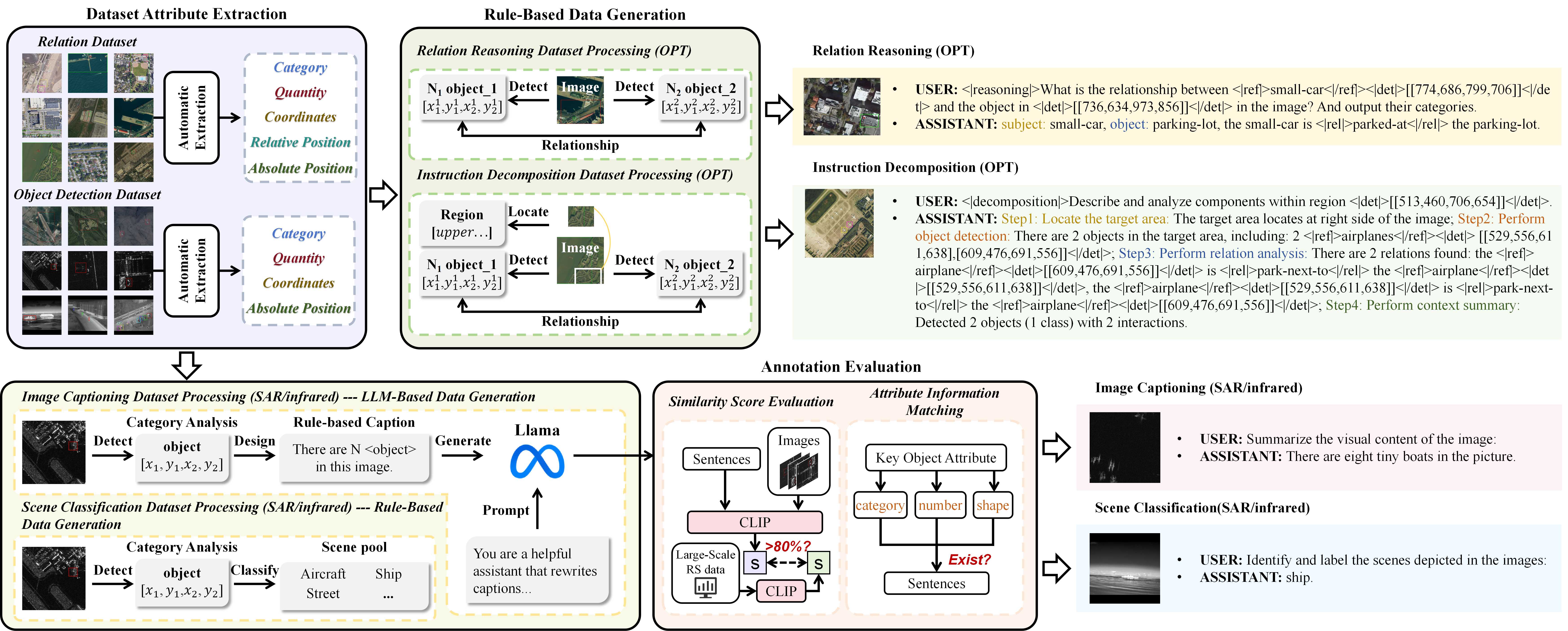}
    \caption{A dataset construction workflow that extracts relevant attributes from existing relation and detection datasets for instruction decomposition, relation reasoning, classification, and captioning tasks.}
\label{fig:process}
\end{figure*} 

\subsection{Dataset Construction for Tasks}

\textcolor{black}{In addition to utilizing publicly available datasets directly, we also construct datasets for complex tasks by processing open-source multi-modal data. The overall workflow is illustrated in \cref{fig:process}. We extract object-level attribute information from optical, SAR, and infrared images and format the data according to the design requirements of each task. For data usage, to ensure consistent spatial representation, all coordinates are normalized to the range $[0, 999]$. The \texttt{[label]} option is used to indicate the modality branch during training and inference, and is presented as clickable buttons in the user interface, with possible values \texttt{opt}, \texttt{sar}, and \texttt{ir}. Details of the data for each task category are provided below.}

\begin{table}[]
\centering
{\color{black}
\caption{The overview of the dataset RS-VL3M, including task types, dataset size, modalities, and image size.\textless x, x\textgreater indicates variable image size, representing the range from minimum to maximum.}
\label{tab1}
\resizebox{0.5\columnwidth}{!}{%
\begin{tabular}{c|rccc}
\hline
Task                                       & Dataset  & Dataset Size & Modality & Image Size \\ \hline
\multirow{1}{*}{Task Scheduling}           &CityNav\cite{lee2024citynav}         & 32,637     & optical     & 384$\times$384             \\ \hline
\multirow{1}{*}{Action Decision}    &SkyAgent-Plan3k\cite{yao2024aeroverse}        &3,000      & optical     & 512$\times$512          \\  
                                           \hline

\multirow{1}{*}{Instruction Decomposition} & ReCon1M-DEC (ours)           & 27,821     & optical      & \textless135, 1000\textgreater  \\ \hline
\multirow{2}{*}{Relation Reasoning}      &  FIT-RS~\cite{luo2024skysensegpt}         & 97,843     & optical     &  512$\times$512      \\  
  &  ReCon1M-REL (ours)        & 125,000      & optical     & \textless135, 1000\textgreater    \\ \hline

\multirow{5}{*}{Object Detection}         &DIOR \cite{li2020object}  &23,463  &optical   & 800$\times$800 \\
         & DOTA\cite{xia2018dota}        &108,047       &optical   & 800$\times$800  \\ 
        & NWPU VHR-10\cite{cheng2014multi}        &800       &optical   & \textless500, 1100\textgreater \\ 
& SARDet-100k \cite{li2024sardet}  & 116,598      & SAR     & \textless190, 1000\textgreater         \\  
& IR-DET (ours)  & 56,353      & infrared     & \textless92, 6912\textgreater        \\  \hline
        

\multirow{6}{*}{Image Captioning} 

&UCM-Captions\cite{qu2016deep}        &10,500     & optical     & 256$\times$256    \\  
&Sydney-Captions\cite{qu2016deep}        &3,065     & optical  & 500$\times$500   \\  
&RSICD\cite{lu2017exploring}        &54,605     & optical     & 224$\times$224  \\  
&NWPU-Captions\cite{cheng2022nwpu}        &157,500     & optical  & 256$\times$256    \\ 
& SAR-CAP (ours)  & 582,990      & SAR     & \textless190, 1000\textgreater   \\  
& IR-CAP (ours)  & 281,765      & infrared     & \textless92, 6912\textgreater   \\  \hline

\multirow{6}{*}{Scene Classification}    &AID\cite{xia2017aid}        &10,000      & optical     & 600$\times$600       \\  
   &  NWPU-RESISC45\cite{cheng2017remote} & 31,500 &optical  & 256$\times$256  \\  
   & WHU-RS19\cite{xia2010structural} &1,005 &optical & 600$\times$600    \\  
   & UCMerced-LandUse \cite{yang2010bag}   & 2,100 &optical & 256$\times$256   \\  
& SAR-CLA (ours)  & 116,597      & SAR     & \textless190, 1000\textgreater        \\  
& IR-CLA (ours)  & 56,353      & infrared     & \textless92, 6912\textgreater       \\  \hline

 \multirow{2}{*}{VQA}    
&RSVQA-HR\cite{lobry2020rsvqa}        &1,066,316     & optical     & 512$\times$512  \\
&RSVQA-LR\cite{lobry2020rsvqa}        &77,232     & optical     & 256$\times$256           \\  
\hline
                                           
\end{tabular}%
}
}
\end{table}

\textbf{Task Scheduling.} This task focuses on trajectory planning for autonomous aerial navigation in complex urban environments based on RS imagery. Given instructions that describe the spatial context surrounding the target location, the model must identify the 3D position of the target based on references to nearby buildings or landmarks, and, given the initial pose of the agent, produce a sequence of 3D waypoints along with corresponding orientation angles. Our task scheduling dataset is derived from CityNav \cite{lee2024citynav} and structured as follows:
\begin{itemize}
    \item Prompt: \textless$|\texttt{navigation}|$\textgreater You need to formulate a flight plan for a quadcopter based on this map, enabling it to fly over all the buildings and reach the destination. The target location is described as follows: \textless$description$\textgreater. The 3D coordinates of the landmark are as follows: $\textless |\texttt{ref}| \textgreater\textless landmark \textgreater \textless |/\texttt{ref}| \textgreater  $ $  \textless |\texttt{pos}| \textgreater[x_1, y_1, z_1] \textless |/\texttt{pos}| \textgreater$. Your starting 3D coordinates and orientation angles are $ \textless |\texttt{pose}| \textgreater[[x_0,y_0,z_0,  $ $  \phi_0, \theta_0, \psi_0]] \textless |/\texttt{pose}| \textgreater$. You need to provide a series of 3D waypoints and attitude angles for the quadcopter to reach the target location.
    \item Response: Step 1: Extract basic information as follows: Target: $\textless target \textgreater$. Landmarks: $\textless landmark \textgreater$. Surroundings: $\textless surrounding \textgreater$.

     Step 2: Get landmarks position: $\textless |\texttt{ref}|\textgreater \textless landmark \textgreater $ $\textless  |/\texttt{ref}| \textgreater \textless |\texttt{pos}| \textgreater[x_1, y_1, z_1] \textless |/\texttt{pos}| \textgreater$. 

      Step 3: Get target position: $\textless |\texttt{ref}|\textgreater \textless target \textgreater \textless  |/\texttt{ref}|$ $ \textgreater \textless |\texttt{pos}| \textgreater[x_2, y_2, z_2] \textless |/\texttt{pos}| \textgreater$. 

      Step 4: Trajectory: $ \textless |\texttt{pose}| \textgreater[[x_0,y_0,z_0, \phi_0, \theta_0, \psi_0],..., $ $ [x_n,y_n,z_n, \phi_n, \theta_n, \psi_n]] \textless |/\texttt{pose}| \textgreater$. 

\end{itemize}

A set of special tokens is embedded in the prompt: $\textless|\texttt{navigation}|\textgreater$ defines the task type. $\textless|\texttt{ref}|\textgreater$ and $\textless|\texttt{/ref}|\textgreater$ denote the name of a landmark, while $\textless|\texttt{pos}|\textgreater$ and $\textless|\texttt{/pos}|\textgreater$ indicate its corresponding coordinates. The trajectory is structured using $\textless| \texttt{pose} |\textgreater$ and $\textless |\texttt{/pose}|\textgreater$ tags, where each pose entry consists of six elements: spatial position $(x, y, z)$ and orientation angles $(\phi, \theta, \psi)$, representing roll, pitch, and yaw respectively. Besides, placeholders such as $\textless landmark \textgreater$, $\textless surrounding \textgreater$, and $\textless target \textgreater$ are used to mark category information (e.g., ``Wellington Road''), while $\textless description \textgreater$ is a placeholder sentence used to describe the spatial surroundings of the target (e.g.,``The row of grayish brown houses on Leslie Road to the left of the gray house at the intersection with Wellington Road'').

\textbf{Action Decision. }
This task focuses on fine-grained action decision planning for aerial agents navigating urban environments, bridging visual-language reasoning with control execution. Given the initial and target 3D poses, along with urban map imagery, the model is required to generate a detailed sequence of instructions. Each instruction corresponds to an interpretable motion step (e.g.,``go straight,'' ``slightly turn right'') and is grounded in observable visual cues such as nearby buildings, colors, shapes, or spatial layouts. Our action decision dataset uses data from SkyAgent-Plan3k \cite{yao2024aeroverse}, as exemplified below:
\begin{itemize}
    \item Prompt: $\textless |\texttt{decision}| \textgreater$How to fly from position$ \textless |  $ $  \texttt{pose}|\textgreater[[x_0,y_0,z_0,\phi_0, \theta_0, \psi_0]] \textless |/\texttt{pose}| \textgreater$ to position$ \textless  $ $   |\texttt{pose}| \textgreater[[x_n,y_n,z_n, \phi_n, \theta_n, \psi_n]] \textless |/\texttt{pose}| \textgreater$, and provide a detailed plan.
    \item Response: Step1: $\textless description  \textgreater$. 
    
    Step2: $\textless description \textgreater$.
    
    \ldots

    Stepn: $\textless description \textgreater$.

\end{itemize}

We introduce a special token $\textless|\texttt{decision}|\textgreater$  to specify task semantics. The usage of $\textless|\texttt{pose}|\textgreater$ and $\textless|\texttt{/pose}|\textgreater$  is consistent with the task scheduling dataset. The starting point appears in the first image, and the destination in the last. The output format requires each scene transition to be annotated with a $\textless description \textgreater$ tag, with each step written from the agent perspective in natural language to support human understanding and decision-making.

\textbf{Instruction Decomposition.} This task aims to improve the capability of the model to execute complex, high-level RS commands by breaking them down into a sequence of fine-grained, interpretable sub-tasks. In our setting, the model receives an image alongside a natural language command that may reference specific regions or semantic goals. The model is expected to decompose this input into a structured multi-step reasoning process, including region localization, object detection, relation analysis, and contextual summarization. Our instruction decomposition dataset builds upon the RS relation comprehension dataset ReCon1M ~\cite{sun2024recon1m}.

\begin{itemize}
    \item Prompt: $\textless|\texttt{decomposition}|\textgreater$Analyze the region $\textless |$ $\texttt{det}|$\textgreater$[[x_1,y_1,x_2,y_2]]$\textless $|/\texttt{det}|\textgreater$ of the image.

    \item Response: Step 1: Locate the target area: The target area locates at the $\textless direction \textgreater$ of the image;
    
    Step 2: Perform object detection: There are $N$ entities in the target area, including: $N_1$ $ \textless  |\texttt{ref}| \textgreater  \textless object \textgreater  $ $  \textless |/\texttt{ref}| \textgreater\textless  |\texttt{det}| \textgreater [[x_1^1,y_1^1,x_2^1,y_2^1]]\textless |/\texttt{det}| \textgreater, ..., N_n  \textless  |  $ $ \texttt{ref}| \textgreater  \textless object \textgreater \textless |/\texttt{ref}| \textgreater \textless  |\texttt{det}| \textgreater[[x_1^n,y_1^n,x_2^n,y_2^n]]\textless |  $ $ /\texttt{det}| \textgreater$;
    
    Step 3: Perform relation analysis: There are $M$ relations found: $\textless  |\texttt{ref}| \textgreater  \textless object \textgreater \textless |/\texttt{ref}| \textgreater \textless  |\texttt{det}| \textgreater[[x_1^1,y_1^1,  $ $ x_2^1,y_2^1]]\textless |/\texttt{det}| \textgreater$ is $\textless |\texttt{rel}| \textgreater  \textless relation \textgreater \textless |/\texttt{rel}| \textgreater$ the $\textless  | \texttt{ref}| \textgreater  \textless object \textgreater \textless |/\texttt{ref}| \textgreater \textless  |\texttt{det}| \textgreater[[x_1^m,y_1^m,x_2^m, $ $ y_2^m]]\textless | /\texttt{det}| \textgreater$;  
    
    Step 4: Perform context summary: $N$ object types with $M$ interactions.
\end{itemize}

As shown above, $\textless direction \textgreater$ has 9 directions, upper, lower, left, right, upper left, upper right, lower left, lower right, and center. $N$ is the total number of detected objects, with $N_1$ to $N_n$ representing category-wise counts. $ \textless  |\texttt{ref}| \textgreater \textless object \textgreater  $ $\textless |/\texttt{ref}| \textgreater\textless  |\texttt{det}| \textgreater [[x_1^1,y_1^1,x_2^1,y_2^1]]\textless |/\texttt{det}| \textgreater$ denotes the specific class of objects with spatial location information, where $[x_1^1,y_1^1,x_2^1,y_2^1]$ denote the top-left and bottom-right corners, respectively. $ \textless  |\texttt{rel}| \textgreater  \textless relation \textgreater \textless |/\texttt{rel}| \textgreater$ denotes the relationship between the two objects.

\textbf{Relation Reasoning.} This task requires the model to infer the categories of the subject and object and to describe the relationship between them, thereby enabling it to recognize objects in the image and reason about their interactions. Our dataset is derived from the RS instruction-tuning dataset FIT-RS~\cite{luo2024skysensegpt} and the RS relation comprehension dataset ReCon1M~\cite{sun2024recon1m}. 



\begin{itemize}
    \item Prompt: $\textless  |\texttt{reasoning}| \textgreater$What is the relationship between $\textless  |\texttt{ref}| \textgreater  \textless object_1 \textgreater\textless |/\texttt{ref}| \textgreater \textless  |\texttt{det}| \textgreater [[x_1^1,y_1^1,  $ $  x_2^1,y_2^1]]\textless |/\texttt{det}| \textgreater$ and the object in $\textless  |\texttt{det}| \textgreater [[x_1^2,y_1^2,x_2^2, $ $ y_2^2]]\textless |/\texttt{det}| \textgreater$ in the image? And output their categories.
    \item Response: subject: $\textless object_1 \textgreater$, object: $\textless object_2 \textgreater$, the $\textless object_1 \textgreater$ is $\textless  |\texttt{rel}| \textgreater  \textless relation \textgreater \textless |/\texttt{rel}| \textgreater$ the $\textless object_2 \textgreater$.
\end{itemize}

Consistent with other tasks, we define a set of special tokens to explicitly mark different semantic components in both the prompt and response.

\textbf{Object Detection.} The optical images are from DIOR \cite{li2020object}, DOTA \cite{xia2018dota}, and NWPU VHR-10 \cite{cheng2014multi}, SAR images are from SARDet-100k\cite{li2024sardet}, and infrared images are from HIT-UAV\cite{HIT-UAV}, Sea shipping\cite{sea-shipping}, Infrared-security\cite{infrared-security}, Aerial-mancer\cite{aerial-mancar}, Double-light-vehicle\cite{double-light-vehicle}, and Oceanic ship\cite {oceanic-ship}. 
\begin{itemize}
    \item Prompt: \textcolor{black}{Detect all objects shown in the remote sensing image and describe using horizontal bounding boxes (HBBs).}
    \item Response: There is/are $ M \textless  |\texttt{ref}| \textgreater \textless object \textgreater \textless |/\texttt{ref}| \textgreater $ $\textless  | \texttt{det}| \textgreater [[x_1^1,y_1^1,x_2^1,y_2^1], ..., [x_1^m,y_1^m,x_2^m,y_2^m]]\textless |/\texttt{det}| \textgreater$ in the image.
    
\end{itemize}

A placeholder $\textless object \textgreater $ is used to represent category names (e.g., ``ship''). The use of coordinates follows the same approach as described previously.

\textbf{Image Captioning.} We construct a multi-modal image captioning benchmark using SAR-CAP, IR-CAP, and various optical captioning datasets. The SAR and infrared data are sourced from the same datasets as the object detection task, while optical data is sourced from public datasets \cite{qu2016deep,lu2017exploring,cheng2022nwpu}.

\textcolor{black}{Optical captions are often richer due to easier manual annotation, typically describing simple scenes with buildings and vegetation. In contrast, SAR and infrared data demand expert analysis. To address this issue, we first generate rule-based captions for objects in the images using standard templates as illustrated in \cref{fig:process}. These captions are then fed as text inputs to Llama3.1 \cite{dubey2024llama}, accompanied by carefully designed prompts that constrain the model to enrich the language while preserving three key attributes: the number, shape, and category of objects. The model produces five distinct textual descriptions for each image. We perform automated validation on the generated texts, which consists of similarity score evaluation and attribute information matching. We compute similarity scores between the generated texts and remote sensing images using RemoteCLIP \cite{liu2024remoteclip}, and compare them with benchmark scores obtained from open-source image captioning datasets. Only texts whose similarity scores exceed 80\% of the benchmark are retained. For attribute consistency, we further validate the number, shape, and category of the objects described in the texts, allowing for synonymous expressions, and retain only the samples that satisfy all these criteria. This approach effectively ensures the accuracy and consistency of the generated descriptions.}

\begin{itemize}
    \item Prompt: Please provide a short depiction of the picture:
    \item Response: A single sailboat is present in the scene.
\end{itemize}

\textbf{Image Classification.} We construct SAR and infrared classification datasets using the same sources as before. For SAR, we adopt the original category definitions, while for infrared data, inconsistent annotations across datasets are addressed by grouping labels into two coarse categories: street scenes and ships. Optical data is sourced from publicly available datasets.
\begin{itemize}
    \item Prompt: Please output the scene corresponding to the image:
    \item Response: aircraft.
\end{itemize}

\textcolor{black}{In our dataset, infrared and SAR images used for detection, captioning, and classification tasks constitute a shared subset, meaning that each image is annotated to support all three tasks. This design enables multi-turn evaluation under consistent visual inputs, while in this study, we validate them as three separate tasks.}

\textbf{VQA.} We use public datasets \cite{ lobry2020rsvqa} as our training and evaluation benchmarks. These datasets are widely adopted as standard benchmarks for evaluating vision-language models.
\begin{itemize}
    \item Prompt: Is a small road present? The answer to this question is 
    \item Response: yes.
\end{itemize}

\section{Method}

\begin{figure*}[tbp]
    \centering
    \includegraphics[width=0.9\linewidth]{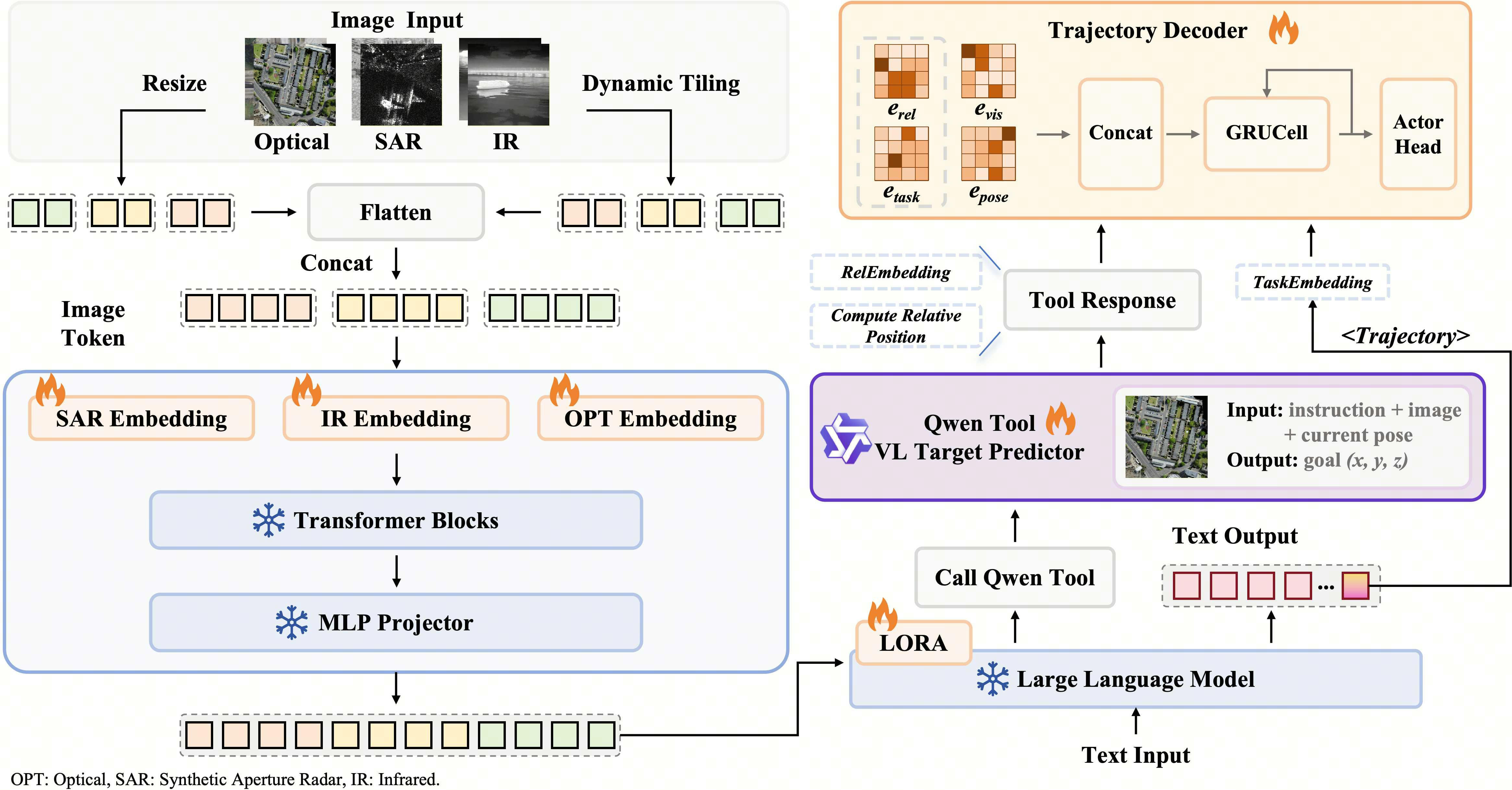}
    \caption{ \textcolor{black}{The overall architecture of RingMo-Agent, which supports multi-platform and multi-modal image inputs for unified perception and reasoning. Image features are encoded by a visual encoder and processed through a frozen MLP and a LoRA-fine-tuned LLM for text generation. For navigation-related tasks, trajectory decoding is further performed with the assistance of an external target prediction tool.}}
\label{fig:architecture}
\end{figure*}

\subsection{Architecture} 
\textcolor{black}{The proposed RingMo-Agent comprises five main components: a visual encoder for extracting visual features, a LLM for processing language information, an MLP projector serving as an intermediate adapter that projects visual features into the semantic space of the language model, a trajectory decoder designed for spatiotemporal modeling, and an external visual grounding tool invoked specifically for absolute coordinate prediction, as illustrated in \cref{fig:architecture}. This unified framework not only processes multi-source data such as optical, infrared, and SAR imagery, but also extends conventional perception tasks to more challenging embodied reasoning tasks, such as autonomous UAV navigation in complex urban scenes.}

\textcolor{black}{\textbf{Visual Encoder.} The visual encoder is based on the SigLIP \cite{zhai2023sigmoid} method, dynamically rescaling input images to multiples of 384 while accommodating RS images with varying aspect ratios. In particular, given an input image of size \( (H \times W \times C ) \in \mathbb{R} \), the image is rescaled to dimensions \( m \times 384, n \times 384, c \), where  \( m, n \in \mathbb{N}, \quad 1 \leq m, n, mn \leq 9\). The values of \( m \) and \( n \) are determined by selecting the smallest multiples of 384 that are greater than or equal to the original image height \( H \) and width \( W \), respectively.  Following this resizing step, the image is partitioned into independent patches of size \( 384 \times 384 \). Additionally, a global thumbnail is generated by directly resizing the image to \( 384 \times 384 \) to provide a coarse global representation. }

\textcolor{black}{Most existing research focuses on single-modality optical learning, while distribution gaps across modalities limit the effectiveness of a unified encoder. To address this issue, we adopt modality-specific embedding layers for optical, SAR, and infrared inputs. Specifically, three parallel patch embedding layers with shared architecture but independent parameters are used to project each modality into a unified 1152-dimensional embedding space, thereby reducing cross-modal interference. To formalize this process, let \( x_{\text{opt}},  x_{\text{ir}}, x_{\text{sar}} \in \mathbb{R}^{H \times W \times C} \) represent the input images from the optical, infrared, and SAR modalities, respectively.} The visual embeddings for each modality are obtained as:
\begin{equation}
z_{\text{opt}} = E_{\text{opt}}(x_{\text{opt}}), \quad z_{\text{ir}} = E_{\text{ir}}(x_{\text{ir}}), \quad z_{\text{sar}} = E_{\text{sar}}(x_{\text{sar}}), 
\end{equation}
where \( E_{\text{opt}} \), \( E_{\text{ir}} \), and \( E_{\text{sar}} \) are the embedding layers for optical, infrared, and SAR data, respectively. These embeddings are then forwarded to the subsequent image encoder for further processing. The frozen visual encoder, denoted as $F_V$, receives an input image $z$ and outputs the corresponding visual features:
\begin{equation}
V = F_V(z), \qquad z =  \{z_{\text{opt}}, z_{\text{ir}}, z_{\text{sar}}\},
\end{equation}
where $V$ represents the extracted visual features.

\textbf{Frozen MLP Projector.} The adapter follows an MLP architecture composed of linear layers.   Following the setup of DeepSeek-VL2  \cite{wu2024deepseek}, the mapping performed by the adapter can be represented by the following computation:
\begin{equation}
Q = F_A(V),
\end{equation}
where, $F_A$ denotes the adapter composed of MLP layers, and $Q$ represents the mapped features.  In $F_A$, a 2$\times$2 shuffling operation is first applied to transform each tile's visual tokens from a 27$\times$27 layout to 14$\times$14. Subsequently, three types of special tokens are inserted into the visual features to serve as positional indicators: one is appended to the end of each row in the global thumbnail tile, another to the end of the last column in the local tiles, and the third is placed between the global and local tiles. Then, as illustrated in \cref{fig:architecture}, the processed embeddings are fed into the LLM as input.















\begin{algorithm}[tbp] 
\small
\caption{\textcolor{black}{Tool-Integrated Trajectory Decoding}} 
\label{algorithm1} 

\textbf{Input: } \textcolor{black}{The hidden representation of the $ \textless \texttt{trajectory} \textgreater $ token $h_{tra}$, the global visual embedding $v$, the absolute goal $g$ predicted by the external tool, the start state $P_{0}$, maximum termination step $T$, distance threshold $p_{th}$, finished flag $P_{flag}$.} 

\textbf{Output: } \textcolor{black}{Generated trajectory $S$.} 

\textcolor{black}{1:\phantom{0} $h_0 \leftarrow $ StartStateEmbedding ($P_0$)} 

\textcolor{black}{2:\phantom{0} $e_{task} \leftarrow $ TaskEmbedding ($h_{tra}$)} 

\textcolor{black}{3:\phantom{0} $e_{vis} \leftarrow $ VisualEmbedding ($v$)} 

\textcolor{black}{4:\phantom{0} $P_{cur} \leftarrow P_0$} 

\textcolor{black}{5:\phantom{0} \textbf{while} $t \leq T$ \textbf{and not} $P_{flag}$ \textbf{do}:} 

\textcolor{black}{6:\phantom{0} \hspace{2em} $e_{pose} \leftarrow $ PoseEmbedding ($P_{cur}$)} 

\textcolor{black}{7:\phantom{0} \hspace{2em} $v_{rel} \leftarrow g - P_{cur}$ } 

\textcolor{black}{8:\phantom{0} \hspace{2em} $e_{rel} \leftarrow $ RelEmbedding ($v_{rel}$)} 

\textcolor{black}{9:\phantom{0} \hspace{2em} $f_t \leftarrow $ Concatenate ($e_{task}, e_{vis}, e_{pose}, e_{rel}$)} 

\textcolor{black}{10: \hspace{2em} $h_t \leftarrow $ GRUCell ($f_t, h_{t-1}$)} 

\textcolor{black}{11: \hspace{2em} $\Delta_t \leftarrow $ DeltaProjection ($h_t$)} 

\textcolor{black}{12: \hspace{2em} $P_{new} \leftarrow P_{cur} + \Delta_t$} 

\textcolor{black}{13: \hspace{2em} $P_{new} \leftarrow $ Clamp ($P_{new}, 0, 1$) \hspace{4.5em} } 

\textcolor{black}{14: \hspace{2em} $S \leftarrow S \cup \{P_{new}\}$} 

\textcolor{black}{15: \hspace{2em} \textbf{if} $||P_{new} - g||_2 < p_{th}$ \textbf{then}} 

\textcolor{black}{16: \hspace{4em} $S \leftarrow S \cup \{g\}$ } 

\textcolor{black}{17: \hspace{4em} $P_{flag} \leftarrow $ \textbf{True}} 

\textcolor{black}{18: \hspace{2em} \textbf{end if}} 

\textcolor{black}{19: \hspace{2em} $P_{cur} \leftarrow P_{new}$} 

\textcolor{black}{20: \hspace{2em} $t \leftarrow t + 1$} 

\textcolor{black}{21: \textbf{end while}} 

\end{algorithm}

\subsection{Architecture and Trajectory Decoding}

\textbf{LoRA-Fine-Tuned Large Language Model.} The core LLM of our framework is built upon the DeepSeekMoE architecture \cite{dai2024deepseekmoe,liu2024deepseek}, which introduces multi-head latent attention by compressing the Key-Value cache into a latent representation to significantly reduce training overhead. Specifically, we utilize the 3B variant of DeepSeek-VL2 \cite{wu2024deepseek} and apply Low-Rank Adaptation (LoRA) \cite{hu2022lora} for parameter-efficient fine-tuning. This is achieved by injecting trainable low-rank matrices into the query and value projection layers of the self-attention blocks, with a rank of 64 and a scaling factor of 16. During fine-tuning, only these newly added parameters are updated. The generation process can be formulated as:
\begin{equation} 
\hat{y} = F_L(Q, x_{txt}), 
\end{equation} 
where $F_L$ denotes the LoRA-adapted LLM, $x_{txt}$ represents the textual input, and $\hat{y}$ is the generated output. 

\textcolor{black}{\textbf{Agent-Driven Navigation with Tool Use.} Conventional vision-language models typically generate text or 2D coordinates via sequential token decoding, which is fundamentally insufficient for modeling continuous, high-dimensional spatial motion. To address this limitation, we formulate our model as an autonomous agent equipped with external tool-use capabilities. When the special $ \textless \texttt{trajectory} \textgreater $ token is triggered during a navigation task, the agent actively invokes an external visual grounding tool to explicitly predict the absolute 3D goal coordinates $g$. If the external tool is unavailable, the system falls back to the language model's own coordinate prediction. For this specific tool, we employ the Qwen3.5-9B-Base model \cite{qwen35blog} and perform full-parameter fine-tuning for 12 epochs, equipping it with precise spatial localization capabilities. Subsequently, the agent's internal trajectory decoder autoregressively generates a continuous dynamic trajectory by integrating this tool-provided goal guidance with the final transformer layer's hidden state. In our data format, these predicted states are represented as the sequence of 3D waypoints and attitude angles enclosed by $\textless|\texttt{pose}|\textgreater\cdots\textless|/\texttt{pose}|\textgreater$.  The complete procedure of this mechanism is detailed in \cref{algorithm1}.}

\textcolor{black}{Within the trajectory decoder, let $h_{\text{tra}}$ denote the embedding vector of the $ \textless \texttt{trajectory} \textgreater $ token extracted from the final layer, and $v$ represent the global visual embedding. We derive the task embedding $e_{task}$ and the visual embedding $e_{vis}$ via linear projections. To provide strong physical priors and explicit directional guidance, at each time step $t$, we compute the relative directional vector $v_{rel}$ between the absolute goal $g$ and the current state $P_{cur}$. The hidden state $h_t$ is updated via:} 
\begin{equation} 
h_t = \text{GRUCell}(f_t, h_{t-1}), 
\end{equation} 
\textcolor{black}{where the input feature $f_t$ is formed by concatenating the task embedding $e_{task}$, the global visual feature $e_{vis}$, the current pose encoded by a linear projection $\phi_{\text{pose}}(P_{cur})$, and the relative direction encoded by a linear projection $\phi_{\text{rel}}(v_{rel})$; $h_0$ is obtained by encoding the starting state $P_0$.}

\textcolor{black}{Next, the decoder predicts the continuous offset $\Delta_t$ to update the current position: }
\begin{equation} 
\Delta_t =  \phi_{\text{delta}}(h_t), 
\end{equation} 
\textcolor{black}{where $\phi_{\text{delta}}$ is an MLP head that predicts the continuous pose increment. To prevent error accumulation and ensure physical feasibility, the updated state $P_{new}$, derived by adding the predicted offset $\Delta_t$ to the current state $P_{cur}$, is explicitly constrained within the normalized valid range of $[0, 1]$ via a boundary clamping operation. The valid predicted state is then appended to the trajectory sequence $S$.}

\textcolor{black}{We use a distance-based stopping criterion for trajectory decoding. A trajectory is marked as completed when the 2D Euclidean distance between the current prediction and the goal is below $p_{\mathrm{th}}=0.035$ in the normalized coordinate space. 
This threshold is stricter than the CityNav success radius, and thus avoids prematurely accepting inaccurate endpoints while tolerating discretization noise. At inference time, decoding stops when all valid samples in the batch have reached the threshold, after which the goal is appended as the final waypoint for endpoint consistency. }

\subsection{\textcolor{black}{Three-Stage Training Methodology}} 
\textcolor{black}{To optimize the proposed architecture, we adopt a three-stage training strategy consisting of RS image-text alignment, multi-task supervised instruction tuning, and a reinforcement learning stage dedicated to trajectory decoding.}

\textcolor{black}{\textbf{Image-Text Alignment.} In the first stage, we adapt the pretrained vision-language model from natural-scene understanding to the RS domain. Since generic pretrained models often struggle with fine-grained RS semantics and may confuse multiple nearby objects, we train the language model on paired RS image-text datasets~\cite{qu2016deep,lu2017exploring,cheng2017remote,xia2010structural,yang2010bag,RSI-CB,RSSCN7,RSD46-WHU,PatternNet,hou2021two,xia2018dota,li2020object} using a generative language modeling objective. 
To improve linguistic diversity while preserving the underlying semantics, GPT-4~\cite{achiam2023gpt} is used to expand the original annotations while keeping the original labels unchanged, after which rule-based checks are applied to verify annotation consistency.} 

\textcolor{black}{\textbf{Supervised Instruction-Tuning.} In the second stage, the model is trained to follow task-specific multimodal instructions.  For navigation, the language generation module and the autoregressive trajectory decoder are jointly optimized.  The decoder is trained with teacher forcing, where the ground-truth goal coordinate and step-wise expert poses provide supervision.  The text objective is the token-level cross-entropy loss $\mathcal{L}_{\mathrm{txt}}$, with additional emphasis on coordinate tokens enclosed by the position markers. The overall supervised objective is:} 
\begin{equation}
\mathcal{L}_{\mathrm{SFT}}
=
\mathcal{L}_{\mathrm{txt}}
+
5\left(\mathcal{L}_{\mathrm{traj}}\right).
\end{equation}
\textcolor{black}{where $\mathcal{L}_{\mathrm{traj}}$ is the masked MSE over valid trajectory states.}

\textcolor{black}{\textbf{Reinforcement Learning for Trajectory Decoding.} To alleviate exposure bias caused by teacher forcing and improve long-horizon robustness, we further optimize the trajectory decoder with GRPO~\cite{shao2024deepseekmath} for a single RL fine-tuning epoch, while keeping all other model components frozen. Given the task-level hidden representation, visual features, start pose, and target goal coordinate, the decoder autoregressively predicts pose increments in a continuous action space. Specifically, each action is sampled from a Gaussian policy parameterized by the decoder output mean and a learnable log standard deviation. During RL sampling, we apply scheduled sampling with a probability of $0.1$, replacing the predicted pose with the corresponding expert pose when available, which stabilizes exploration and reduces early-stage drift.}

\textcolor{black}{For a generated trajectory $S$, the trajectory-level reward is defined as:}
\begin{equation}
R(S) = - (\lambda_{\text{dist}} \mathcal{D}_{\text{final}} + \lambda_{\text{smooth}} \mathcal{P}_{\text{smooth}}) + R_{\text{success}},
\end{equation}
\textcolor{black}{where $\mathcal{D}_{\text{final}}$ denotes the final Euclidean distance to the target goal, $\mathcal{P}_{\text{smooth}}$ is the mean acceleration penalty for encouraging smooth motion, and $R_{\text{success}}$ is a sparse bonus assigned when the predicted trajectory terminates within the target threshold. In practice, we set $\lambda_{\text{dist}}=2.0$, $\lambda_{\text{smooth}}=0.1$, and $R_{\text{success}}=5.0$.}

\textcolor{black}{Instead of learning a value function, GRPO computes group-normalized advantages from $G$ sampled trajectories for each instruction and optimizes the policy via a clipped surrogate objective:}
\begin{equation}
\mathcal{L}_{\text{GRPO}} = - \frac{1}{G} \sum_{i=1}^{G} \min \left( \rho_i(\theta) A_i,\; \mathrm{clip}(\rho_i(\theta), 1-\epsilon, 1+\epsilon) A_i \right),
\end{equation}
\textcolor{black}{where $\rho_i(\theta)=\frac{\pi_\theta(a|s)}{\pi_{\theta_{\text{old}}}(a|s)}$ is the policy ratio and $A_i$ is the group-relative advantage. To prevent unstable exploration and preserve reasonable flight dynamics, we further regularize RL training with a supervised trajectory loss, and the final objective is written as}
\begin{equation}
\mathcal{L} = \mathcal{L}_{\text{GRPO}} + \lambda_{\text{SFT}} \mathcal{L}_{\text{traj}},
\end{equation}
\textcolor{black}{where $\mathcal{L}_{\text{traj}}$ denotes the point-wise trajectory regression loss and $\lambda_{\text{SFT}}=0.5$. This hybrid training strategy improves goal-reaching performance while maintaining smooth and physically plausible navigation trajectories.}

\section{Experiments}

This section details the training procedure and evaluates the instruction-tuned model on both perception tasks (detection, VQA, classification, captioning) and reasoning tasks (relation reasoning, task scheduling, instruction decomposition, action decision), using both qualitative and quantitative analyses. 

In the reported results, FT denotes the fine-tuned results, while ZS refers to the zero-shot results. \textbf{Bold} indicates the best performance, and \uline{underline} denotes the second-best. 

\subsection{Experimental Settings}
During vision-language generation stage, we train our model on 8 NVIDIA A100 GPUs (80GB) using the AdamW optimizer. The learning rate follows a linear warmup followed by a cosine decay schedule, with an initial learning rate of 1e-4, a warmup learning rate of 1e-6, and a minimum learning rate of 1e-5. We apply a weight decay of 0.05 for regularization and train the model on over 500,000 samples for 10 epochs, with the image size fixed at 384.

During instruction-tuning stage, RingMo-Agent adopts a linear warmup followed by a cosine decay learning rate schedule. The training utilizes the AdamW optimizer, with an initial learning rate of 1e-6, a warmup starting from 1e-8, and a minimum learning rate decaying to 0. To further enhance generalization and mitigate overfitting, a weight decay of 0.05 is applied throughout the optimization process.  In this stage, we impose no restriction on image resolution and retain the original dimensions of each sample. The model is also trained for 10 epochs. The combined training, validation, and test sets contain over 3 million samples, collectively referred to as RS-VL3M. 

\subsection{Task Scheduling}
\textbf{Dataset.} \textcolor{black}{We use the CityNav dataset \cite{lee2024citynav}, which comprises 34 diverse urban scene environments. The dataset encompasses a wide range of target object categories.}

\begin{table*}[htbp]
\centering
\textcolor{black}{
\caption{Evaluation results on the CityNav dataset.}
\label{tab:citynav}
\renewcommand{\arraystretch}{1.1}
\resizebox{0.9\textwidth}{!}{%
\begin{tabular}{lcccc|cccc|cccc}
\toprule
\multicolumn{1}{l}{\multirow{2}{*}{Model}} & \multicolumn{4}{c}{Validation Seen} & \multicolumn{4}{c}{Validation Unseen} & \multicolumn{4}{c}{Test Unseen} \\ \cline{2-13} 
\multicolumn{1}{c}{}   & NE $\downarrow$  &  SR $\uparrow$  & OSR $\uparrow$  &  SPL $\uparrow$ & NE $\downarrow$  &  SR $\uparrow$  & OSR $\uparrow$  &  SPL $\uparrow$ & NE $\downarrow$  &  SR $\uparrow$  & OSR $\uparrow$  &  SPL $\uparrow$ \\ \midrule
\textit{\textbf{Specialist models}} \\
Seq2Seq (FT) \cite{anderson2018vision} &257.1	&1.81	&7.89	&1.58	&317.4	&0.79	&8.82	&0.61	&245.3	&1.50	&8.34	&1.30
\\
CMA (FT) \cite{anderson2018vision}  &240.8	&0.95	&9.42	&0.92	&268.8	&0.65	&7.86	&0.63	&252.6	&0.82	&9.70	&0.79
\\
AerialVLN + GSM (FT) \cite{liu2023aerialvln} &\uline{56.6}	&\uline{10.16}	&\uline{22.20}	&\uline{7.89}	&\uline{72.7}	&\uline{6.35}	&\uline{15.24}	&\uline{5.06}	&\uline{85.1}	&\uline{6.72}	&\uline{18.21}	&\uline{5.16}\\ \midrule
\textit{\textbf{RSVLMs}} \\
 \cellcolor{gray!10}RingMo-Agent (FT)     & \cellcolor{gray!10}\textcolor{black}{\textbf{35.7}} & \cellcolor{gray!10}\textcolor{black}{\textbf{45.81}}  & \cellcolor{gray!10}\textcolor{black}{\textbf{57.07}} & \cellcolor{gray!10}\textcolor{black}{\textbf{45.59}} 
 & \cellcolor{gray!10}\textcolor{black}{\textbf{61.6}} & \cellcolor{gray!10}\textcolor{black}{\textbf{16.99}}  & \cellcolor{gray!10}\textcolor{black}{\textbf{30.11}} & \cellcolor{gray!10}\textcolor{black}{\textbf{16.66}} 
 & \cellcolor{gray!10}\textcolor{black}{\textbf{64.2}} & \cellcolor{gray!10}\textcolor{black}{\textbf{21.15}}  & \cellcolor{gray!10}\textcolor{black}{\textbf{33.79}} & \cellcolor{gray!10}\textcolor{black}{\textbf{20.92}}
     \\ 
     \bottomrule
\end{tabular}%
}
}
\end{table*}

\textbf{Metrics.} \textcolor{black}{We evaluate on all three test splits. Following CityNav \cite{lee2024citynav}, four metrics are used to assess the accuracy of the generated trajectories: Navigation Error (NE), Success Rate (SR), Oracle Success Rate (OSR), and Success weighted by Path Length (SPL).} 

\textbf{Results.} \textcolor{black}{As shown in \cref{tab:citynav}, RingMo-Agent achieves competitive results on both the validation and test sets, consistently outperforming the fine-tuned Seq2Seq~\cite{anderson2018vision}, CMA~\cite{liu2023aerialvln}, and AerialVLN~\cite{lee2024citynav}, despite the use of map encoding and multi-view features in AerialVLN. In particular, RingMo-Agent obtains lower NE and substantially higher SR, OSR, and SPL across
all splits, demonstrating stronger goal-reaching accuracy and long-horizon navigation robustness. These improvements mainly benefit from the effective external tool trained to predict the absolute target position, which provides explicit goal guidance, as well as the additional reinforcement learning stage introduced to optimize the trajectory decoder under its rollout distribution. In contrast, current general-purpose vision-language models (VLMs) and remote sensing vision-language models (RSVLMs) still struggle to complete this complex spatial navigation task.}

\subsection{Action Decision}

\textbf{Dataset.}  We evaluate this task on the SkyAgent-Plan3k \cite{yao2024aeroverse} dataset, which encompasses four distinct urban scene types: Shanghai, Shenzhen, Campus, and Residence.

\textbf{Metrics.} To assess the quality of the generated step-by-step action plans, we adopt BLEU and SPICE as evaluation metrics. 

\begin{table*}[htbp]
\centering
\textcolor{black}{
\caption{Evaluation results on the SkyAgent-Plan3k dataset.}
\label{tab:plan3k}
\renewcommand{\arraystretch}{1.2}
\resizebox{0.9\textwidth}{!}{%
\begin{tabular}{lcc|cc|cc|cc} 
\toprule
\multicolumn{1}{l}{\multirow{2}{*}{Model}} & \multicolumn{2}{c}{ShangHai} & \multicolumn{2}{c}{ShenZhen} & \multicolumn{2}{c}{Campus} & \multicolumn{2}{c}{Residence} \\ \cline{2-9} 
\multicolumn{1}{c}{}   & BLEU-1 $\uparrow$   &  SPICE $\uparrow$   & BLEU-1 $\uparrow$  &  SPICE $\uparrow$  & BLEU-1$\uparrow$  &  SPICE $\uparrow$  & BLEU-1 $\uparrow$  &  SPICE $\uparrow$  \\ \midrule
\textit{\textbf{VLMs}} \\
InstructBLIP-Flan-T5-XXL (ZS)  \cite{dai2023instructblip}  &0.04 &5.79 &0.03 &4.23 &0.95 &4.58 &0.14 &7.20\\
InstructBLIP-Vicuna-7B (ZS)  \cite{dai2023instructblip}  &0.15&0.68&0.04&3.23&0.74&1.21&0.18&1.86\\
InstructBLIP-Vicuna-13B (ZS)  \cite{dai2023instructblip}  &0.12&1.14&0.34&2.18&0.63&0.46&0.07&1.51\\
     MiniGPT-4 (ZS) \cite{zhu2023minigpt} &  12.88 & 4.03 & 13.96  & 4.51   & 16.89 & 4.25    & 15.85   & 4.51            \\
     GPT-4o (ZS) \cite{hurst2024gpt}  & 10.64  & 5.12 & 11.52 & 5.20   & 11.67  & 5.03 & 11.83 & 4.83    \\ 
     Qwen-VL (ZS) \cite{bai2023qwenvlversatilevisionlanguagemodel}&15.74&5.41&15.36&5.06&12.18&3.10&10.01&3.75\\
     \midrule
\textit{\textbf{RSVLMs}} \\
SkyAgent (FT) \cite{yao2024aeroverse} & \textbf{40.87} & \uline{26.53} & \uline{36.05} & \uline{24.04} & \uline{34.03} & \uline{22.57} & \uline{42.56} & \uline{24.29}  \\
 \cellcolor{gray!10}RingMo-Agent (FT)     & \cellcolor{gray!10}\uline{37.33}  &\cellcolor{gray!10}\textbf{31.68}  &\cellcolor{gray!10}\textbf{37.76} &\cellcolor{gray!10}\textbf{31.85}    &\cellcolor{gray!10}\textbf{42.79} &\cellcolor{gray!10}\textbf{28.45}  &\cellcolor{gray!10}\textbf{42.78}  &\cellcolor{gray!10}\textbf{37.68}  
     \\ \bottomrule
\end{tabular}%
}
}
\end{table*}

 \textbf{Results.}  \textcolor{black}{As shown in \cref{tab:plan3k}, our method demonstrates clear advantages over the baseline approaches across all scene categories. Existing RSVLMs have yet to explore their reasoning capabilities in 3D space guided by multi-image inputs. These results validate the effectiveness of our model in bridging the gap between visual-language reasoning and actionable control.}

\subsection{Relation Reasoning}

\textbf{Dataset.} We conduct evaluation experiments on two relation reasoning datasets: the FIT-RS dataset \cite{luo2024skysensegpt} and our self-constructed ReCon1M-REL dataset. The FIT-RS dataset includes 54 relation categories, with all images cropped to a fixed resolution of 512$\times$512. In contrast, ReCon1M-REL retains 59 relation categories and adopts variable resolutions. 

\textbf{Metrics.} We use F1-score to assess this task.

\begin{table}[h]
\centering
\textcolor{black}{
\caption{Evaluation results on the FIT-RS and ReCon1M-REL datasets.}
\label{tab:relation}
\renewcommand{\arraystretch}{1.2}
\resizebox{0.45\columnwidth}{!}{%
\begin{tabular}{lcc}
\toprule
\multicolumn{1}{l}{\multirow{2}{*}{Model}} & \multicolumn{1}{c}{FIT-RS} & \multicolumn{1}{c}{ReCon1M-REL}\\ \cline{2-3} 
\multicolumn{1}{c}{}   & F1-score $\uparrow$      &  F1-score    $\uparrow$   \\ \midrule
\textit{\textbf{VLMs}} \\
   MiniGPT-v2 (ZS) \cite{chen2023minigpt}        &     0.00    &0.00\\ 
   DeepSeek-VL2 (ZS) \cite{wu2024deepseek}  &0.06 & 0.30\\
   Kosmos-2 (ZS) \cite{peng2023kosmos}  &0.00 & 0.15\\
   Shikra (ZS) \cite{chen2023shikra}  &0.52 & 0.65\\
   Qwen3-VL (ZS) \cite{bai2025qwen3vltechnicalreport} &0.00 &\uline{4.56} \\
Gemini 2.5 (ZS) \cite{comanici2025gemini}&0.41 &1.36\\
GPT-4o (ZS) \cite{hurst2024gpt}  & 3.32& 3.04\\
   \midrule
\textit{\textbf{RSVLMs}}  \\
SkySenseGPT (FT) \cite{luo2024skysensegpt}     &     \uline{74.33}      &  -  \\
\cellcolor{gray!10}\cellcolor{gray!10}RingMo-Agent (FT)    &  \cellcolor{gray!10}\textbf{75.34}      &    \cellcolor{gray!10}\textbf{90.23} \cellcolor{gray!10} \\ \bottomrule
\end{tabular}%
}
}
\end{table}

\textbf{Results.} As presented in \cref{tab:relation}, RingMo-Agent demonstrates superior performance, achieving an accuracy of 75.34\%, which slightly surpasses the 74.33\% obtained by SkySenseGPT \cite{luo2024skysensegpt}. On ReCon1M-REL dataset, we achieve the F1-score of 90.23\%. \textcolor{black}{In contrast, state-of-the-art VLMs \cite{bai2025qwen3vltechnicalreport, comanici2025gemini, hurst2024gpt} are still unable to effectively perform such tasks, even when provided with more explicit prompts. This limitation primarily stems from the lack of targeted training that emphasizes relation reasoning among objects in wide-field images, which consequently leads to substantially degraded performance.}

\subsection{Instruction Decomposition}

\textbf{Dataset.} We evaluate our model on the self-constructed ReCon1M-DEC dataset.

\textbf{Metrics.} \textcolor{black}{To comprehensively assess performance, we evaluate the models using mAP@50, which measures mean Average Precision at an Intersection-over-Union threshold of 0.5, and the F1-score.}

\begin{table}[h]
\centering
\textcolor{black}{
\caption{Evaluation results on the ReCon1M-DEC dataset.}
\label{tab:recon1m-dec}
\renewcommand{\arraystretch}{1.2}
\resizebox{0.45\columnwidth}{!}{%
\begin{tabular}{lccc}
\toprule
Model & mAP@50 $\uparrow$ & F1-Score $\uparrow$ \\ \midrule
   \textit{\textbf{VLMs}} \\
        MiniGPT-v2 (ZS) \cite{chen2023minigpt}  &     0.00    &0.00 \\
   DeepSeek-VL2 (ZS) \cite{wu2024deepseek}   &0.00&0.00\\
      Qwen3-VL (ZS) \cite{bai2025qwen3vltechnicalreport}&0.28 &0.03\\
Gemini 2.5 (ZS) \cite{comanici2025gemini}&\uline{0.50} &0.07\\
GPT-4o (ZS) \cite{hurst2024gpt}  &0.14 &\uline{0.13} \\
   \midrule
   \textit{\textbf{RSVLMs}} \\
  \cellcolor{gray!10}RingMo-Agent (FT)   &\cellcolor{gray!10}\textbf{24.20} &\cellcolor{gray!10}\textbf{32.85}    \\ \bottomrule
\end{tabular}%
}
}
\end{table}

\textbf{Results.}  \textcolor{black}{ \Cref{tab:recon1m-dec} reports object recognition accuracy and relation reasoning performance within the target region. We employ carefully designed prompts to evaluate advanced VLMs as baselines. As shown, existing VLMs exhibit notably low accuracy, largely due to insufficient training emphasis on object categories, spatial locations, and inter-object relationships in wide-area scenes. Consequently, when multiple objects are present, these models typically predict only one or two instances, failing to reflect the true object multiplicity. In contrast, RingMo-Agent effectively handles the challenging ReCon1M-DEC dataset by capturing fine-grained features of small objects and producing correct one-to-one relational predictions, achieving an mAP@50 of 24.20\% for object detection and an F1-score of 32.85\% for intra-region relation detection.}

\subsection{Image Captioning}

\textbf{Dataset.} We conduct evaluation separately on optical, SAR, and infrared modalities. The optical datasets utilize UCM-Captions \cite{qu2016deep } and NWPU-Captions \cite{cheng2022nwpu}, while the SAR and infrared modalities are tested on our self-constructed datasets.

\textbf{Metrics.} The performance is reported using BLEU, METEOR, ROUGE-L, and CIDEr metrics.

\begin{table*}[htbp]
\centering
\textcolor{black}{
\caption{Evaluation results on the SAR-CAP and IR-CAP datasets.}
\label{tab:sar-ir-cap}
\renewcommand{\arraystretch}{1.2}
\resizebox{\textwidth}{!}{%
\begin{tabular}{lcccccc|cccccc} 
\toprule
\multicolumn{1}{l}{\multirow{2}{*}{Model}} & \multicolumn{6}{c}{SAR-CAP} & \multicolumn{6}{c}{IR-CAP} \\ \cline{2-13} 
\multicolumn{1}{c}{}   & BLEU-1 $\uparrow$ & BLEU-2 $\uparrow$& BLEU-3 $\uparrow$ & BLEU-4 $\uparrow$& METEOR $\uparrow$ & ROUGE-L $\uparrow$  & BLEU-1 $\uparrow$ & BLEU-2 $\uparrow$& BLEU-3  $\uparrow$& BLEU-4 $\uparrow$& METEOR $\uparrow$ & ROUGE-L $\uparrow$ \\ \midrule
\textit{\textbf{VLMs}} \\
MiniGPT-v2 (ZS) \cite{chen2023minigpt} & 7.00&3.64 & 1.59& 0.60&7.67 & 9.25&5.65 &3.35 & 1.92 &0.98&7.62&8.67\\
DeepSeek-VL2 (ZS) \cite{wu2024deepseek}  &12.52 &5.88 & 1.88 & 0.60 &10.65 &14.10  & 13.95& 7.57 &3.47  & 1.43  &12.48  &15.12 \\
Qwen3-VL (ZS) \cite{bai2025qwen3vltechnicalreport} & 12.51&5.70 &1.43 &0.42 &11.74 &12.85 & 7.62 & 4.30 &1.91 &0.82 &8.43 &9.90 \\
Gemini 2.5 (ZS) \cite{comanici2025gemini}& 13.00& 5.43&1.81 &0.77 &12.46 &13.23 & 17.70& 7.76& 3.17& 1.32& 14.14& 17.99 \\
GPT-4o (ZS) \cite{hurst2024gpt} &\uline{19.93}&\uline{9.76}&\uline{3.14}&\uline{1.15}&\uline{13.52}&\uline{19.24} &\uline{28.05} &\uline{14.43} &\uline{6.15} & \uline{2.41} & \uline{16.23} & \uline{26.76}\\
   \midrule
\textit{\textbf{RSVLMs}} \\
 \cellcolor{gray!10}RingMo-Agent (FT)  & \cellcolor{gray!10}\textbf{55.93} &\cellcolor{gray!10}\textbf{44.49}  &\cellcolor{gray!10}\textbf{33.57} &\cellcolor{gray!10}\textbf{23.94}     &\cellcolor{gray!10}\textbf{25.06}  &\cellcolor{gray!10}\textbf{51.12} &\cellcolor{gray!10}\textbf{56.84}&\cellcolor{gray!10}\textbf{40.45}&\cellcolor{gray!10}\textbf{29.17}&\cellcolor{gray!10}\textbf{21.50}&\cellcolor{gray!10}\textbf{26.15} &\cellcolor{gray!10}\textbf{43.13}
     \\ \bottomrule
\end{tabular}%
}
}
\end{table*}

\begin{table}[h]
\centering
\textcolor{black}{
\caption{Evaluation results on the UCM Dataset.}
\label{tab:ucm_caption}
\resizebox{0.5\columnwidth}{!}{%
\begin{tabular}{lcccc}
\toprule
Model & BLEU-4 $\uparrow$& METEOR $\uparrow$ & ROUGE-L $\uparrow$& CIDEr $\uparrow$\\ \midrule
\textit{\textbf{Specialist models}} &              &        &         &  \\
 SAA  (FT) \cite{lu2019sound}   &  64.77      &  38.59      &    69.42     &   294.51    \\
   Post-processing (FT) \cite{hoxha2023improving}   &  62.62      &     40.80         &    74.06     &   309.64    \\ \midrule
\textit{\textbf{VLMs}} &              &        &         &  \\
    MiniGPT-4 (ZS) \cite{zhu2023minigpt} &    18.10   &     33.36       &   41.37      &   0.03    \\
  Shikra (ZS) \cite{chen2023shikra}   &    33.98   &      32.56      &    56.73     &    56.69   \\ 
   MiniGPT-v2 (ZS) \cite{chen2023minigpt}   & 36.16&  32.41 & 56.57 & 60.66 \\ 
   DeepSeek-VL2 (ZS) \cite{wu2024deepseek}  &5.44 & 15.95&27.22 & 26.42\\
Qwen3-VL (ZS) \cite{bai2025qwen3vltechnicalreport}&1.82 &15.73 &24.92 &8.85\\
Gemini 2.5 (ZS) \cite{comanici2025gemini}&2.78 &17.05 &25.96 &11.74\\
GPT-4o (ZS) \cite{hurst2024gpt}  &3.33&15.37&26.31&21.13 \\
   \midrule
\textit{\textbf{RSVLMs}} &              &        &         &  \\
    RSGPT (FT) \cite{hu2023rsgpt} &      65.74  &       42.21     &  78.34       &  333.23     \\
   RingMoGPT (FT) \cite{wang2024ringmogpt}  &   -  &   \uline{49.90}        &   83.29     &    \uline{359.32}   \\ 
   RS-CapRet (FT) \cite{silva2024large} &67.00 &47.20 &81.70 &354.80 \\   
   RS-LLaVA (FT) \cite{bazi2024rs} & 72.84 & 47.98 & \uline{85.17} & 349.43 \\
   SkyEyeGPT (FT)  \cite{zhan2025skyeyegpt} &  \textbf{78.41} &   46.24        &   79.49    &    236.75   \\ 
   \cellcolor{gray!10}RingMo-Agent (FT)   &   \cellcolor{gray!10}\uline{77.63}  &     \cellcolor{gray!10}\textbf{51.79}     &   \cellcolor{gray!10}\textbf{85.51}    &  \cellcolor{gray!10}\textbf{373.68}   \\ \bottomrule
\end{tabular}%
}
}
\end{table} 

\begin{table}[h]
\centering
\textcolor{black}{
\caption{Zero-shot results on the NWPU-Captions dataset.}
\label{tab:caption_nwpu}
\resizebox{0.5\columnwidth}{!}{%
\begin{tabular}{lccc}
\hline
\toprule
Model & METEOR $\uparrow$ & ROUGE-L $\uparrow$& CIDEr $\uparrow$\\ \midrule
\textit{\textbf{VLMs}} &              \\
Qwen-VL\cite{bai2023qwenvlversatilevisionlanguagemodel}          & 12.60           & 26.24              & 24.64           \\
MiniGPT-v2\cite{chen2023minigpt}     & 13.70 & 26.60 & 19.70         \\
LLaVA\cite{liu2024visual}            & 13.70          & 28.80          & 32.60           \\ 
DeepSeek-VL2 \cite{wu2024deepseek}  & 14.88&19.56 & 3.18\\
   Qwen3-VL \cite{bai2025qwen3vltechnicalreport} &15.92&23.91&7.72\\
   Gemini 2.5 \cite{comanici2025gemini}&16.17&24.00&8.40\\
   GPT-4o \cite{hurst2024gpt}  &15.93&27.33&15.33\\
   \midrule
\textit{\textbf{RSVLMs}} &                &  \\
RingMoGPT \cite{wang2024ringmogpt} & \textbf{20.90}         &  \textbf{37.50}         & \textbf{74.70} \\ \cellcolor{gray!10}RingMo-Agent &   \cellcolor{gray!10}\uline{18.81}  &     \cellcolor{gray!10}\uline{34.11}     &   \cellcolor{gray!10}\uline{48.39}       \\ \bottomrule
\end{tabular}}
}
\end{table}

\textbf{Results.} \textcolor{black}{As shown in \cref{tab:ucm_caption} and \cref{tab:caption_nwpu}, RingMo-Agent consistently outperforms both specialized models and VLMs on the UCM-Captions \cite{qu2016deep }  and NWPU-Captions \cite{cheng2022nwpu} optical image captioning datasets. On UCM-Captions, it achieves a METEOR score of 51.79, a ROUGE-L score of 85.51, and a CIDEr score of 373.68, surpassing existing RSVLMs. Furthermore, on SAR and infrared captioning benchmarks (\cref{tab:sar-ir-cap}), RingMo-Agent demonstrates a clear advantage over advanced VLMs such as Qwen3-VL \cite{bai2025qwen3vltechnicalreport}. This performance gap arises because these models are insufficiently trained on multi-modal RS data; in particular, on SAR imagery, they tend to misclassify objects such as ships and aircraft as stars.}

\subsection{VQA}

\textbf{Dataset.} We report the fine-tuned result on RSVQA-LR \cite{lobry2020rsvqa} test set and zero-shot result on RSVQA-HR \cite{lobry2020rsvqa}. These benchmarks assess the model’s ability to understand object types, quantities, and spatial locations.

\textbf{Metrics.} We compute the accuracy for each question type as well as the overall average accuracy.

\begin{table}[h]
\centering
\textcolor{black}{
\caption{Evaluation results on the RSVQA-LR Dataset.}
\label{tab:RSVQA-LR}
\resizebox{0.5\columnwidth}{!}{%
\begin{tabular}{lccc}
\toprule
Model & Pre. Acc  $\uparrow$& Comp. Acc $\uparrow$ & Avg. Acc  $\uparrow$ \\ \midrule
\textit{\textbf{Specialist models}} &              &        &   \\
EasyToHard (FT) \cite{yuan2022easy} &90.66 &87.49 &89.08 \\
Bi-Modal (FT) \cite{bazi2022bi} &91.06 &91.16 &91.11 \\
SHRNet (FT) \cite{zhang2023spatial} &91.03 &90.48 &90.76
 \\ \midrule
\textit{\textbf{VLMs}} &              &        &      \\
 MiniGPT-4 (ZS) \cite{zhu2023minigpt} &43.86 &57.55 &50.71   \\
Shikra (ZS) \cite{chen2023shikra}  &46.47 &60.31 &53.39\\
Qwen-VL (ZS) \cite{bai2023qwenvlversatilevisionlanguagemodel} &38.57 &67.59 &53.08 \\
MiniGPT-v2 (ZS)  \cite{chen2023minigpt} &49.85 &63.09 &56.47 \\
InstructBLIP (ZS)  \cite{dai2023instructblip}  &48.83 &65.92 &57.38\\
mPLUG-Owl2 (ZS)  \cite{ye2024mplug} &74.04& 63.69 &68.87 \\
LLaVA-1.5 (ZS)   \cite{liu2024improved}   &55.46 &68.20 &61.83 \\ 
DeepSeek-VL2 (ZS) \cite{wu2024deepseek}  & 55.33 & 66.87 &61.10 \\
Qwen3-VL (ZS) \cite{bai2025qwen3vltechnicalreport} &66.53 &77.49 &72.01\\
Gemini 2.5 (ZS) \cite{comanici2025gemini}&67.99 &72.77 &70.38\\
GPT-4o (ZS) \cite{hurst2024gpt} &69.20 &63.14&66.17\\
\midrule
\textit{\textbf{RSVLMs}} &              &        &        \\
 RSGPT (FT)  \cite{hu2023rsgpt}  &91.17 &\textbf{91.70} &\uline{91.44} \\
 Geochat (FT)     \cite{kuckreja2024geochat} &91.09 &90.33 &90.71 \\
LHRS-Bot (FT)      \cite{muhtar2024lhrs} &89.07& 88.51 &88.79\\
H$^2$RSVLM (FT)     \cite{pang2024h2rsvlm}  &89.58 &89.79 &89.69 \\
RS-LLaVA  (FT)    \cite{bazi2024rs}   &\uline{92.80} &\uline{91.33} &\textbf{92.07} \\
SkyEyeGPT  (FT)  \cite{zhan2025skyeyegpt} &88.93 &88.63 &88.78\\
   \cellcolor{gray!10}RingMo-Agent   (FT)   &   \cellcolor{gray!10}\textbf{93.10}  &     \cellcolor{gray!10}87.50     &   \cellcolor{gray!10}90.30  \\ \bottomrule
\end{tabular}%
}
}
\end{table}

\begin{table}[h]
\centering
\caption{Zero-shot results on the RSVQA-HR dataset.}
\label{tab:RSVQA-HR}
\resizebox{0.5\columnwidth}{!}{%
\begin{tabular}{lccc}
\toprule
Model & Pre. Acc  $\uparrow$& Comp. Acc $\uparrow$ & Avg. Acc  $\uparrow$ \\ \midrule
\textit{\textbf{VLMs}} &              &        &  \\
\textcolor{black}{Qwen-VL} \cite{bai2023qwenvlversatilevisionlanguagemodel} &66.44 &60.41 &63.43 \\
LLaVA-v1.5 \cite{liu2024visual}  &69.83 & 67.29 &68.56\\
MiniGPT-v2 \cite{chen2023minigpt} &40.79 &50.91 &45.85 \\
\midrule
\textit{\textbf{RSVLMs}} &              &        &        \\ 
EarthGPT \cite{zhang2024earthgpt} & 62.77 & 79.53 &71.15 \\
 Geochat  \cite{kuckreja2024geochat} &58.45 &83.19 &70.82 \\
H$^2$RSVLM  \cite{pang2024h2rsvlm}  &65.00 &83.70 &74.35 \\
SkySenseGPT  \cite{luo2024skysensegpt} &\uline{69.14} &\textbf{84.14} &\uline{76.64}\\
   \cellcolor{gray!10}RingMo-Agent    &\cellcolor{gray!10}\textbf{75.24}  &\cellcolor{gray!10}\uline{83.92}   &\cellcolor{gray!10}\textbf{79.58}  \\ \bottomrule
\end{tabular}%
}
\end{table}

\textbf{Results.} \textcolor{black}{\cref{tab:RSVQA-LR} presents the results on RSVQA-LR, focusing on two question types: Presence and Comparison. Compared with other methods, we achieve 93.10\% accuracy on presence questions, surpassing RS-LLaVA \cite{bazi2024rs} by 0.3\%. \cref{tab:RSVQA-HR} shows that we achieve accuracies of 75.24\% on presence questions, 83.92\% on comparison questions, and an overall average accuracy of 79.58\%.  Moreover, our model consistently outperforms VLMs on both datasets, further demonstrating its suitability for RS applications.}

\subsection{Classification}

\textbf{Dataset.} We report results on the AID \cite{xia2017aid} and NWPU-RESISC45 \cite{cheng2017remote} datasets, along with our self-constructed IR-CLA and SAR-CLA datasets, and provide zero-shot results on UCMerced-LandUse \cite{yang2010bag} and WHU-RS19 \cite{xia2010structural}.

\textbf{Metrics.} \textcolor{black}{For optical datasets, following the approach of GeoChat \cite{kuckreja2024geochat}, we use all categories as candidate answers to prevent synonymous outputs. For infrared and SAR datasets, all categories are provided as candidate answers for VLMs.}

\begin{table}[h]
\centering
\textcolor{black}{
\caption{Evaluation results on the AID and NWPU-RESISC45 datasets.}
\label{tab:aid}
\resizebox{0.45\columnwidth}{!}{%
\begin{tabular}{lcc}
\toprule
Model & AID (Acc  $\uparrow$) & NWPU-RESISC45 (Acc $\uparrow$)\\ \midrule
\textit{\textbf{Specialist models}} &              &      \\
LSENet (FT) \cite{bi2021local} &\textbf{94.41} &93.34\\
SeCo-ResNet-50 (FT) \cite{manas2021seasonal} &\uline{93.47} &92.91\\
  \midrule
\textit{\textbf{VLMs}} &              &         \\
DeepSeek-VL2 (ZS) \cite{wu2024deepseek}  & 35.63&17.60 \\
Qwen3-VL (ZS) \cite{bai2025qwen3vltechnicalreport} &70.71 &71.01\\
Gemini 2.5 (ZS) \cite{comanici2025gemini}& 37.97&68.58\\
GPT-4o (ZS) \cite{hurst2024gpt} & 58.33 &50.52\\
 \midrule
\textit{\textbf{RSVLMs }} &              &         \\
EarthGPT (FT) \cite{zhang2024earthgpt} &- &\uline{93.84} \\
\cellcolor{gray!10}RingMo-Agent (FT)  &   \cellcolor{gray!10}92.43 &     \cellcolor{gray!10}\textbf{93.96} \\ \bottomrule
\end{tabular}%
}
}
\end{table}

\begin{table}[h]
\centering
\textcolor{black}{
\caption{Zero-shot results on the WHU-RS19 and UCMerced-LandUse datasets.}
\label{tab:whu}
\resizebox{0.45\columnwidth}{!}{%
\begin{tabular}{lcc}
\toprule
Model &  WHU-RS19 (Acc  $\uparrow$) & UCMerced-LandUse (Acc $\uparrow$)\\ \midrule
\textit{\textbf{VLMs}} &              &         \\
Qwen-VL \cite{bai2023qwenvlversatilevisionlanguagemodel} &83.51 &62.90 \\
MiniGPT-v2 \cite{chen2023minigpt} &62.08 &4.76 \\
LLaVA\cite{liu2024visual}  &74.52 &68.00 \\
CLIP\cite{radford2021learning}  &87.50 &– \\
DeepSeek-VL2 \cite{wu2024deepseek}  &69.05 & 32.00\\
Qwen3-VL \cite{bai2025qwen3vltechnicalreport} &82.39&77.81\\
Gemini 2.5 \cite{comanici2025gemini}&47.56 &65.10\\
GPT-4o \cite{hurst2024gpt} &85.27&43.52\\
\midrule
\textit{\textbf{RSVLMs}} &              &         \\
RemoteCLIP \cite{liu2024remoteclip} &94.66 &- \\
GeoChat \cite{kuckreja2024geochat} &86.47 &84.43 \\
RingMoGPT \cite{wang2024ringmogpt} &\textbf{97.71} &\uline{86.48}\\
SkySenseGPT \cite{luo2024skysensegpt} &\uline{97.02} &-\\
\cellcolor{gray!10}RingMo-Agent  &   \cellcolor{gray!10}95.28  &     \cellcolor{gray!10}\textbf{86.52} \\ \bottomrule
\end{tabular}%
}}
\end{table}

\begin{table}[h]
\centering
\textcolor{black}{
\caption{Evaluation results on the SAR-CLA and IR-CLA datasets. }
\label{tab:sar-ir-cla}
\renewcommand{\arraystretch}{1.2}
\resizebox{0.4\columnwidth}{!}{%
\begin{tabular}{lc|c} 
\toprule
\multicolumn{1}{l}{\multirow{2}{*}{Model}} & \multicolumn{1}{c}{SAR-CLA} & \multicolumn{1}{c}{IR-CLA} \\ \cline{2-3} 
\multicolumn{1}{c}{}   & Acc  $\uparrow$  & Acc  $\uparrow$  \\ 
   \midrule
\textit{\textbf{VLMs}} \\
DeepSeek-VL2 (ZS) \cite{wu2024deepseek}  & 5.22 &95.56 \\
   Qwen3-VL (ZS) \cite{bai2025qwen3vltechnicalreport} &29.38 & 98.21\\
   Gemini 2.5 (ZS) \cite{comanici2025gemini}&\uline{37.89} &97.55\\
   GPT-4o (ZS) \cite{hurst2024gpt}  &36.74 & \uline{98.42}\\
   \midrule
\textit{\textbf{RSVLMs}} \\
 \cellcolor{gray!10}RingMo-Agent (FT)  & \cellcolor{gray!10}\textbf{92.67}  &\cellcolor{gray!10}\textbf{99.45}  
     \\ \bottomrule
\end{tabular}%
}}
\end{table}

\textbf{Results.} \textcolor{black}{As shown in \cref{tab:aid}, RingMo-Agent achieves strong performance on optical tasks, reaching 93.96\% accuracy on NWPU-RESISC45. On the zero-shot set UCMerced-LandUse, we achieve 86.52\%, surpassing other RSVLMs, as shown in \cref{tab:whu}. This improvement is attributed to the robust generalization ability of the model.} 

\textcolor{black}{For the infrared classification dataset IR-CLA in \cref{tab:sar-ir-cla}, which has only two possible classes, VLMs generally produce correct predictions following prompting, though their performance remains slightly below that of RingMo-Agent. In contrast, for the SAR-CLA dataset in \cref{tab:sar-ir-cla}, some images contain multiple class labels, and the combination of modality-specific challenges and multi-label prediction complexity limits the performance of advanced VLMs that were not trained for such tasks. These results demonstrate the effectiveness of our model in RS multi-modal classification.}

\subsection{Object Detection}

\textbf{Dataset.} RingMo-Agent supports object detection across three modalities: optical, SAR, and infrared. SAR results are reported on SARDet-100k \cite{li2024sardet}, and infrared results on our IR-DET dataset.

\textbf{Metrics.}  We report both per-class and overall detection performance using mAP@50.

\begin{table}[h]
\centering
\caption{Evaluation results on the SARDet-100k dataset.}
\label{tab:det-sar}
\resizebox{0.8\columnwidth}{!}{%
\begin{tabular}{lccccccc}
\toprule
Model & Ship & Aircraft & Bridge & Harbor & Car  & Tank & mAP@50 $\uparrow$ \\ \midrule

\textit{\textbf{VLMs}} \\
MiniGPT-v2 (ZS) \cite{chen2023minigpt} & 0.00 & 0.00 & 0.00 & 0.00 & 0.00 & 0.00 & 0.00\\
DeepSeek-VL2 (ZS) \cite{wu2024deepseek}  & 0.00 & 0.00 & 0.00 & 0.00 & 0.00 & 0.00 & 0.00\\
Kosmos-2 (ZS) \cite{peng2023kosmos} &0.27&\uline{0.17}&0.00&0.00&0.04&0.00&0.08\\
shikra (ZS) \cite{chen2023shikra} &0.00 &0.00  &0.00&0.00&0.00 &0.00 &0.00\\
Qwen3-VL (ZS) \cite{bai2025qwen3vltechnicalreport} & 0.89 & 0.09 &0.00 & 0.00& \uline{0.33} & 0.00 & 0.22 \\
Gemini 2.5 (ZS) \cite{comanici2025gemini}&\uline{1.46} &0.14 &0.00 &0.00 &0.00 &0.00 &\uline{0.27} \\
GPT-4o (ZS) \cite{hurst2024gpt} &0.03&0.00&0.00&0.00&0.00&0.00&0.01\\
   \midrule
\textit{\textbf{RSVLMs}} \\
   \cellcolor{gray!10}RingMo-Agent (FT)   &   \cellcolor{gray!10}\textbf{74.43}  &   \cellcolor{gray!10}\textbf{52.93}   &\cellcolor{gray!10}\textbf{49.90}       &  \cellcolor{gray!10}\textbf{62.30}  &  \cellcolor{gray!10}\textbf{63.37}     &  \cellcolor{gray!10}\textbf{20.10} & \cellcolor{gray!10}\textbf{53.84} \\ \bottomrule
\end{tabular}%
}
\end{table}

\begin{table*}[h]
\centering
\caption{Evaluation results on the IR-DET dataset. }
\label{tab:det-IR}
\renewcommand{\arraystretch}{1.2}
\resizebox{1.0\textwidth}{!}{%
\begin{tabular}{lccccccccccc|cccccc} 
\toprule
\multicolumn{1}{l}{\multirow{2}{*}{Model}} & \multicolumn{11}{c}{Fine-Grained Categories} & \multicolumn{5}{c}{Coarse-Grained Categories} &\multicolumn{1}{c}{\multirow{2}{*}{mAP@50 $\uparrow$}}\\ \cline{2-17} 
\multicolumn{1}{c}{}   & Bulk Carrier & Bus & Canoe & Container Ship & Fishing Boat & Liner & Nonmotor Vehicle & Other Vehicle & Sailboat & Truck & Warship & Bike & Cyclist & People & Ship & Vehicle  
\\ 
   \midrule
\textit{\textbf{VLMs}} \\
   
MiniGPT-v2 (ZS) \cite{chen2023minigpt} & 0.00 & 0.00 & 0.00 & 0.00 & 0.00 & 0.00 & 0.00 & 0.00 & 0.00 & 0.00 & 0.00 & 0.00 & 0.00 & 0.00 & 0.00 & 0.00 & 0.00  \\ 
DeepSeek-VL2 (ZS) \cite{wu2024deepseek}  & 0.00 & 0.00 & 0.00 & 0.00 & 0.00 & 0.00 & 0.00 & 0.00 & 0.00 & 0.00 & 0.00 & 0.00 & 0.00 & 0.00 & 0.00 & 0.00 & 0.00  \\
Kosmos-2 (ZS) \cite{peng2023kosmos} & 0.00 & 8.75 & 0.00 & 0.00 & 0.00 & 0.00 & 0.00 & 0.00 & 3.96 & 1.78 & 0.00 & 0.00 & 0.00 & 1.23 & 2.13 & 7.76& 1.60\\
shikra (ZS) \cite{chen2023shikra} &0.00 &1.81&0.00 &0.00 &0.00 &0.00 &\uline{0.31} &0.00 &0.91 &0.52 &0.00 &0.00 &0.00 &0.95 &0.28 &0.73 &0.34\\
Qwen3-VL (ZS) \cite{bai2025qwen3vltechnicalreport} &0.00&13.34&0.00&0.00&0.00&0.90&0.15&0.00&\uline{8.85}&\uline{16.06}&13.01 &0.00&0.00&1.35&\uline{6.17}&19.11&4.93 \\
Gemini 2.5 (ZS) \cite{comanici2025gemini}&0.00 &\uline{28.79}&0.00 &0.00 &0.00 &\uline{1.18} &0.14& 0.00&2.18&13.95&\uline{24.82}&0.00&\uline{0.50}&\uline{8.26}&2.27 &\uline{22.70}&\uline{6.55} \\
GPT-4o (ZS) \cite{hurst2024gpt} &0.00& 0.25 &0.00&0.00&0.00&0.00&0.00&0.00&0.02&0.02&0.00&0.00&0.00&0.00&0.00&0.03&0.02 \\
   \midrule
\textit{\textbf{RSVLMs}} \\
 \cellcolor{gray!10}RingMo-Agent (FT) &\cellcolor{gray!10}\textbf{70.63} &\cellcolor{gray!10}\textbf{21.25}  &\cellcolor{gray!10}\textbf{74.43}  &\cellcolor{gray!10}\textbf{59.77}  &\cellcolor{gray!10}\textbf{91.47} &\cellcolor{gray!10}\textbf{87.03} &\cellcolor{gray!10}\textbf{78.11}  &\cellcolor{gray!10}\textbf{62.71}  &\cellcolor{gray!10}\textbf{70.80}  &\cellcolor{gray!10}\textbf{39.61}  &\cellcolor{gray!10}\textbf{89.66} &\cellcolor{gray!10}\textbf{35.95} &\cellcolor{gray!10}\textbf{16.61} &\cellcolor{gray!10}\textbf{53.67}   &\cellcolor{gray!10}\textbf{42.52} &\cellcolor{gray!10}\textbf{63.83}  &\cellcolor{gray!10}\textbf{59.88}  
     \\ \bottomrule
\end{tabular}%
}
\end{table*}

 \textbf{Results.} \textcolor{black}{As shown in \cref{tab:det-sar} and \cref{tab:det-IR}, RingMo-Agent significantly outperforms existing models in SAR and infrared object detection. Although advanced generalist VLMs are used as baselines and demonstrate strong detection ability on natural images, they struggle with specific modalities in RS. For instance, Qwen3-VL \cite{bai2025qwen3vltechnicalreport} often misinterprets SAR images as dark or gray sky scenes, reflecting a misunderstanding of SAR imaging characteristics. On the IR-DET dataset, these models can partially recognize common objects such as vehicles and trucks, yet achieve near-zero performance on most categories.  This degradation mainly results from the severe modality shift between natural RGB imagery and non-optical RS data, such as SAR and infrared images, whose physical imaging mechanisms and visual patterns fundamentally differ from those seen during VLM pre-training.}

\subsection{Ablation Study}

To assess the effectiveness of key design components within RingMo-Agent, we conduct ablation experiments from three perspectives: (1) the impact of the image-text alignment  stage training paradigm, (2) the role of modality-specific embedding layers, and (3) the function of the trajectory decoder.

\begin{table*}[h]
\centering
\caption{\textcolor{black}{Ablation results on the impact of image-text alignment stage training for reasoning tasks.}}
\label{tab:Ablation_1_reasoning}
\renewcommand{\arraystretch}{1.1}
\resizebox{0.8\textwidth}{!}{%
\begin{tabular}{c|cc|cc} 
\toprule
    \multicolumn{1}{c}{\multirow{2}{*}{\textcolor{black}{Alignment Stage}}} & \multicolumn{2}{c}{Relation Reasoning} & \multicolumn{2}{c}{Instruction Decomposition} \\ \cline{2-5} 
\multicolumn{1}{c}{}  & FIT-RS (F1-score $\uparrow$)  & ReCon1M-REL (F1-score $\uparrow$) &ReCon1M-DEC (mAP@50 $\uparrow$)  & ReCon1M-DEC (F1-score $\uparrow$) \\ \midrule

\textcolor{black}{$\times$}  &72.82 &87.11 &16.72 & 20.32  \\
 \cellcolor{gray!10}\textcolor{black}{$\checkmark$}   & \cellcolor{gray!10}\textbf{75.34} &\cellcolor{gray!10}\textbf{90.23}  &\cellcolor{gray!10}\textbf{24.20} &\cellcolor{gray!10}\textbf{32.85}  
     \\ \bottomrule
\end{tabular}%
}
\end{table*}

\begin{table*}[h]
\centering
\caption{\textcolor{black}{Ablation results on the impact of image-text alignment stage training for perception tasks.}}
\label{tab:Ablation_1_perception}
\renewcommand{\arraystretch}{1.2}
\resizebox{\textwidth}{!}{%
\begin{tabular}{c|cc|ccc|c} 
\toprule
    \multicolumn{1}{c}{\multirow{2}{*}{\textcolor{black}{Alignment Stage}}} & \multicolumn{2}{c}{Object Detection} & \multicolumn{3}{c}{Image Captioning} & \multicolumn{1}{c}{VQA} \\ \cline{2-7} 
\multicolumn{1}{c}{}  & SARDet-100k (mAP@50 $\uparrow$) & IR-DET (mAP@50 $\uparrow$)& UCM (BLEU-4 $\uparrow$) & SAR-CAP (BLEU-1 $\uparrow$) & IR-CAP (BLEU-1 $\uparrow$) & RSVQA-LR (Avg. Acc $\uparrow$) \\ \midrule

\textcolor{black}{$\times$}  &50.12 & 51.56 & 76.12 & 50.12 & 49.16 & 89.96\\
 \cellcolor{gray!10}\textcolor{black}{$\checkmark$}  & \cellcolor{gray!10}\textbf{53.84} &\cellcolor{gray!10}\textbf{59.88}  &\cellcolor{gray!10}\textbf{77.63} &\cellcolor{gray!10}\textbf{55.93} &\cellcolor{gray!10}\textbf{56.84} &\cellcolor{gray!10}\textbf{90.30} 
     \\ \bottomrule
\end{tabular}%
}
\end{table*}

\textbf{Image-Text Alignment Stage Training.} We pretrain RingMo-Agent on a large-scale RS image-text dataset to enhance image-text generation. To assess the impact, we compare it with a baseline initialized from DeepSeek-VL2 without pretraining, using identical datasets and training strategies. As shown in \cref{tab:Ablation_1_reasoning} and \cref{tab:Ablation_1_perception}, the image-text alignment stage training consistently improves performance on both reasoning and perception tasks. This gain stems from the ability to capture RS-specific features such as resolution, viewpoint, and spectral differences, thereby boosting generalization.

\begin{table*}[h]
\centering
\caption{Ablation results on the impact of modality-specific embedding laters, where MSEL denotes modality-specific embedding layers.}
\label{tab:Ablation_2}
\renewcommand{\arraystretch}{1.2}
\resizebox{\textwidth}{!}{%
\begin{tabular}{c|cc|ccc|cc} 
\toprule
    \multicolumn{1}{c}{\multirow{2}{*}{MSEL}}& \multicolumn{2}{c}{Object Detection} & \multicolumn{3}{c}{Image Captioning} & \multicolumn{2}{c}{Image Classification} \\ \cline{2-8} 
 \multicolumn{1}{c}{}  & SARDet-100k (mAP@50 $\uparrow$) & IR-DET (mAP@50 $\uparrow$)& UCM (BLEU-4 $\uparrow$) & SAR-CAP (BLEU-1 $\uparrow$) & IR-CAP (BLEU-1 $\uparrow$) & SAR-CLA (Acc $\uparrow$) & IR-CLA (Acc $\uparrow$)\\ \midrule

$\times$ &48.10 & 51.23 & 74.11 & 46.32 & 51.08 & 88.17 & 96.60 \\
\cellcolor{gray!10}$\checkmark$ & \cellcolor{gray!10}\textbf{53.84} &\cellcolor{gray!10}\textbf{59.88}  &\cellcolor{gray!10}\textbf{77.63} &\cellcolor{gray!10}\textbf{55.93} &\cellcolor{gray!10}\textbf{56.84} &\cellcolor{gray!10}\textbf{92.67} &\cellcolor{gray!10}\textbf{99.45} 
     \\ \bottomrule
\end{tabular}%
}
\end{table*}

\textbf{Modality-Specific Embedding Layers.}  We fine-tune separate embedding layers for optical, SAR, and infrared modalities, guided by modality labels during training and inference. Only the embedding layers are updated, while the rest of the visual encoder remains frozen. As shown in \cref{tab:Ablation_2}, freezing these layers and removing modality-specific design degrades performance, highlighting the importance of dedicated extractors. Despite sharing identical structures, separate embeddings better capture modality-specific characteristics, avoiding feature degradation caused by forcing heterogeneous data through shared filters.

\begin{table}[h]
\centering
\caption{\textcolor{black}{Ablation results on the impact of external tool integration and RL optimization on embodied navigation tasks.}}
\label{tab:Ablation_3}
\renewcommand{\arraystretch}{1.2}%
\begin{tabular}{c|c|cccc}
\toprule
\multicolumn{1}{c}{\multirow{2}{*}{Tools Used}} & \multicolumn{1}{c}{\multirow{2}{*}{RL Used}} & \multicolumn{4}{c}{Validation Unseen}\\ \cline{3-6} 
\multicolumn{1}{c}{} &\multicolumn{1}{c}{}   & NE $\downarrow$  &  SR $\uparrow$  & OSR $\uparrow$  &  SPL $\uparrow$ \\ \midrule

\textcolor{black}{$\times$}   &\textcolor{black}{$\times$}    &\textcolor{black}{156.0} &\textcolor{black}{4.91}  &\textcolor{black}{17.22} &\textcolor{black}{4.33} \\ 
\textcolor{black}{$\checkmark$}   &\textcolor{black}{$\times$}    &\textcolor{black}{61.9} &\textcolor{black}{16.81} &\textcolor{black}{29.58} &\textcolor{black}{16.49}\\
\cellcolor{gray!10}\textcolor{black}{$\checkmark$}   &\cellcolor{gray!10}\textcolor{black}{$\checkmark$} 
 & \cellcolor{gray!10}\textcolor{black}{\textbf{61.6}} & \cellcolor{gray!10}\textcolor{black}{\textbf{16.99}}  & \cellcolor{gray!10}\textcolor{black}{\textbf{30.11}} & \cellcolor{gray!10}\textcolor{black}{\textbf{16.66}} 
     \\   \bottomrule
\end{tabular}%

\end{table}
 
\textcolor{black}{\textbf{Tool Integration and RL Optimization.} To empower the model as an autonomous entity capable of executing long-horizon embodied navigation tasks, we integrate visual grounding tools and employ a RL optimized trajectory decoding mechanism. In \cref{tab:Ablation_3}, we evaluate the individual contributions of these two components. Without external tools and RL, the target points are directly predicted by the LLM itself in an autoregressive manner. The results on the CityNav dataset \cite{lee2024citynav} demonstrate that the integration of visual grounding tools significantly enhances navigation performance, drastically reducing the NE from 156.0 to 61.9 and improving the SR from 4.91\% to 16.81\%. Furthermore, applying RL to optimize the trajectory decoding process yields the best overall performance across all metrics in unseen scenarios. This confirms the necessity and effectiveness of combining external tool-use capabilities with RL optimization for complex embodied navigation.}

\subsection{Visualization}
\begin{figure*}[h]
    \centering
    \includegraphics[width=0.9\linewidth]{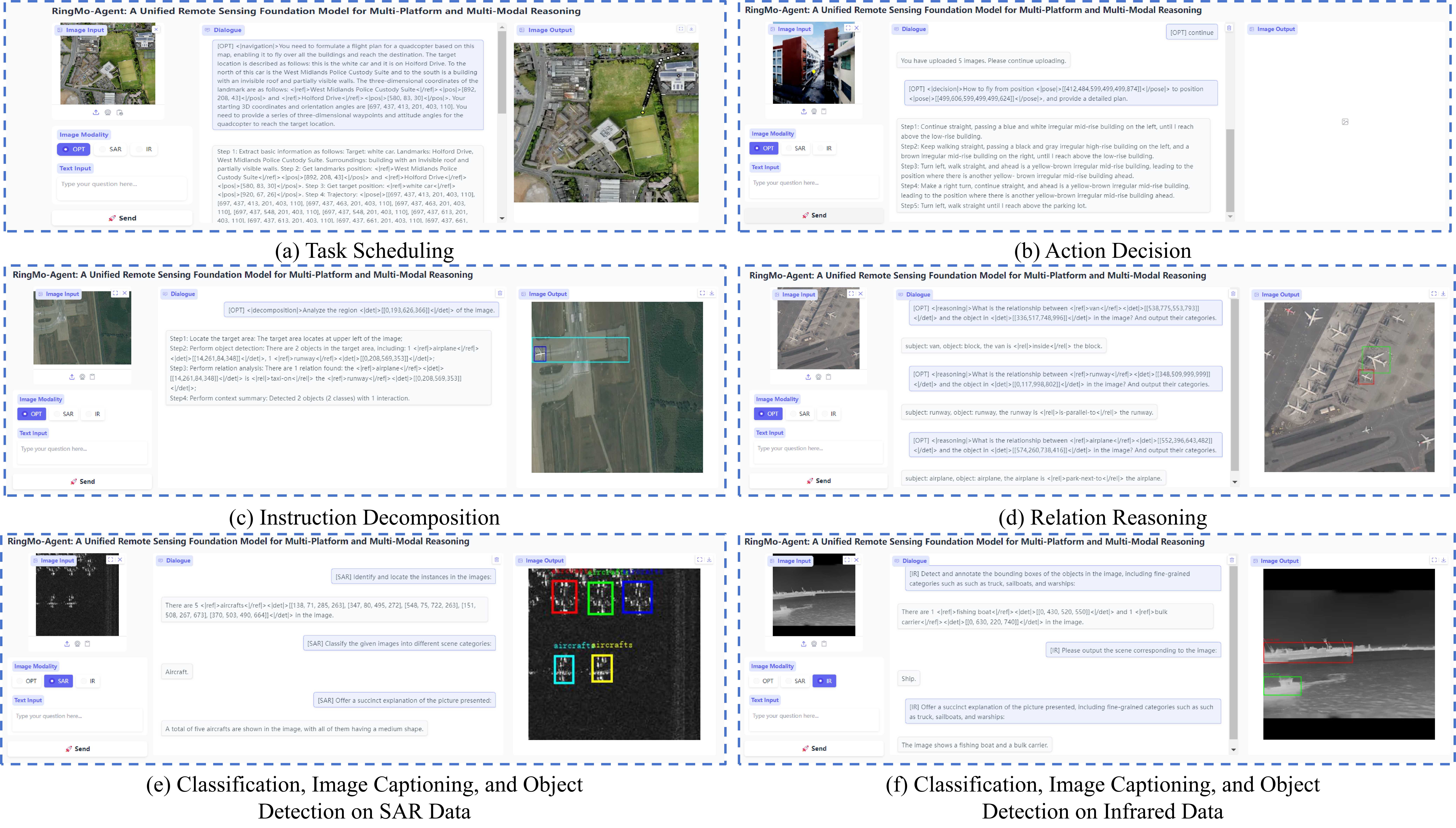}
    \caption{The visual results presented by RingMo-Agent in response to user questions about image content.}
\label{fig:alltasks}
\end{figure*}

To provide a clearer understanding of the model behavior, we present visualizations of representative results in this work, as shown in \cref{fig:alltasks}. These include task scheduling (a), action decision (b), instruction decomposition (c), relation reasoning (d), as well as classification, captioning, and detection on SAR and infrared data (e, f).

\section{Conclusion}

In this work, we propose RingMo-Agent, a unified foundation model  tailored for RS data across multiple platforms and modalities.  It is capable of handling images with diverse viewing angles and sensing characteristics, and performing complex reasoning tasks. The model benefits from several key supports: a large-scale dataset RS-VL3M comprising over 3 million image-text pairs;  the construction of separated embedding layers to address distribution differences across modalities and isolate the feature extraction process; \textcolor{black}{and an agent-driven navigation mechanism that combines external tool use with a token-based high-dimensional hidden state decoding mechanism, which significantly expands the task coverage.} We envision that future RS agents will further advance toward deeper logical reasoning abilities, thereby supporting real-world applications such as damage assessment and decision-making in emergency scenarios like disaster response.

\normalem
\bibliographystyle{IEEEtran}  
\bibliography{references}

\end{document}